\documentclass[letterpaper]{article} 
\usepackage{aaai25}  
\usepackage{times}  
\usepackage{helvet}  
\usepackage{courier}  
\usepackage[hyphens]{url}  
\usepackage{graphicx} 
\usepackage{natbib}  
\usepackage{caption} 
\usepackage{algorithm}
\usepackage{algorithmic}
\usepackage{algorithm}
\usepackage{amsmath}
\usepackage{adjustbox}
\usepackage{multirow}
\usepackage{amssymb}
\usepackage{anyfontsize}
\usepackage{blindtext}
\usepackage{multirow}
\usepackage{amssymb}
\usepackage{pifont}
\usepackage{booktabs} 
\usepackage{color, colortbl}
\usepackage[dvipsnames]{xcolor}
\usepackage[dvipsnames]{xcolor}
\usepackage[labelformat=simple]{subcaption}

\usepackage{algorithm}
\usepackage{algorithmic}
\newtheorem{definition}{Definition} 
\definecolor{Gray}{gray}{0.9}
\definecolor{LightCyan}{rgb}{0.88, 1, 1}

\usepackage{newfloat}
\usepackage{listings}
\DeclareCaptionStyle{ruled}{labelfont=normalfont,labelsep=colon,strut=off} 
\floatstyle{ruled}
\newfloat{listing}{tb}{lst}{}
\floatname{listing}{Listing}
\title{Pre-training Graph Neural Networks on Molecules by Using Subgraph-Conditioned Graph Information Bottleneck}

\author{
    Van Thuy Hoang, O-Joun Lee\thanks{Corresponding author: O-Joun Lee (Tel.: +82-2-2164-5516)}
}
\affiliations{

    Department of Artificial Intelligence, The Catholic University of Korea \\
    {\{hoangvanthuy90, ojlee\}@catholic.ac.kr}
}

\usepackage{bibentry}

\begin{document}

\maketitle


\begin{abstract}
This study aims to build a pre-trained Graph Neural Network (GNN) model on molecules without human annotations or prior knowledge.
Although various attempts have been proposed to overcome limitations in acquiring labeled molecules, the previous pre-training methods still rely on semantic subgraphs, i.e., functional groups.
Only focusing on the functional groups could overlook the graph-level distinctions.
The key challenge to build a pre-trained GNN on molecules is how to (1) generate well-distinguished graph-level representations and (2) automatically discover the functional groups without prior knowledge.
To solve it, we propose a novel Subgraph-conditioned Graph Information Bottleneck, named S-CGIB, for pre-training GNNs to recognize core subgraphs (graph cores) and significant subgraphs.
The main idea is that the graph cores contain compressed and sufficient information that could generate well-distinguished graph-level representations and reconstruct the input graph conditioned on significant subgraphs across molecules under the S-CGIB principle.
To discover significant subgraphs without prior knowledge about functional groups, we propose generating a set of functional group candidates, i.e., ego networks, and using an attention-based interaction between the graph core and the candidates.
Despite being identified from self-supervised learning, our learned subgraphs match the real-world functional groups.
Extensive experiments on molecule datasets across various domains demonstrate the superiority of S-CGIB.

\end{abstract}

%

\section{Introduction}

Graph Neural Networks (GNNs) have recently emerged in computational chemistry, offering powerful tools for predicting molecular properties \cite{DBLP:journals/corr/GilmerSRVD17,DBLP:journals/jcamd/KearnesMBPR16}.
While GNNs have shown remarkable performance in molecular property prediction, their effectiveness depends on the availability of abundant labeled molecules for model training \cite{DBLP:conf/kdd/HaoLHWHLCL20,s23084168}.


Recently, pre-training strategies have offered considerable potential in overcoming the challenges of the scarcity of labeled molecular data \cite{DBLP:conf/nips/RongBXX0HH20,DBLP:conf/nips/LuongS23}.
The existing pre-training strategies can be categorized into three primary groups: node-level pre-training, contrastive learning, and subgraph-level pre-training.
Node-level pre-training mainly focuses on node-level prediction, e.g., node attribute reconstruction or edge prediction, which may not fully leverage the high-order structure of molecules, i.e., functional groups \cite{DBLP:conf/nips/RongBXX0HH20}.
The second strategy is contrastive learning, which focuses on learning representations by contrasting multiple views of molecules based on random or heuristic augmentations \cite{DBLP:conf/icml/YouCSW21,DBLP:conf/icml/XuWNGT21}. 
More recently, subgraph-level strategies focus on semantic subgraphs, which can capture both local and global structural patterns by identifying functional groups \cite{DBLP:conf/aaai/Subramonian21,DBLP:conf/nips/ZhangLWLL21,DBLP:conf/nips/RongBXX0HH20,DBLP:conf/nips/LiuSZZKWC23,inae2024motifaware}.
The main idea is to use human annotations or prior knowledge to extract the semantic subgraphs, e.g., frequent subgraphs across molecules, to enhance recognizing significant substructures and molecular property prediction \cite{DBLP:conf/nips/LuongS23,degen2008art}.
To sum up, most recent pre-training strategies aggregate information from node-level or subgraph-level to generate graph-level representations.





However, two challenges limit the existing pre-training strategies on molecules.
\textit{\textbf{First}}, the existing strategies lack the ability to generate well-distinguished graph-level representations.
Most node-level strategies mainly focus on the local structure and then adopt a pooling function, e.g., mean, max, or sum, to aggregate information, resulting in poor-distinguished graph-level representations as information from noisy and redundant nodes can be aggregated to form graph-level representations \cite{DBLP:conf/iclr/HuLGZLPL20}.
Besides, while subgraph-level strategies \cite{DBLP:conf/aaai/Subramonian21,DBLP:conf/nips/ZhangLWLL21} can capture specific subgraphs at multiple scales, they could overlook the entire graph-level distinctions \cite{DBLP:conf/nips/RongBXX0HH20,inae2024motifaware}.
That is, subgraph-level representations based on discrete pre-defined patterns, e.g., frequent subgraphs, ignore global interactions between important nodes that derive the molecule's entire structure.
For contrastive learning strategies, applying augmentation schemes, e.g., edge perturbation, could potentially disrupt the structures and properties of molecules \cite{DBLP:conf/aaai/Lee0P22}.
\textit{\textbf{Second}}, it is challenging for subgraph-based strategies to cover all possible functional groups given in diverse molecule datasets.
To capture functional groups, recent subgraph-level strategies create dictionaries based on pre-defined rules, e.g., counting the discrete occurrences of subgraphs across molecules to decompose molecules into a bag of subgraphs, commonly prioritizing subgraphs with larger sizes and frequent occurrences \cite{DBLP:conf/nips/LuongS23,DBLP:conf/nips/Kong0T022}.
The prioritization and static dictionaries could limit the model's ability to capture new and uncommon functional groups.

In this paper, we overcome the above challenge of pre-training strategies by considering how to (1) generate well-separated graph-level representations and (2) automatically capture significant subgraphs without explicit annotations or prior knowledge.
The fundamental idea behind our strategy is that, across the chemical domain, molecules share universal core subgraphs that can combine with specific significant subgraphs to robust representations of molecules.

To generate well-distinguished graph-level representations, this initiates a problem in recognizing a set of important nodes (graph cores) that can allow robust and well-separated representations \cite{DBLP:conf/iclr/HuLGZLPL20}.
We interpret this problem as the Graph Information Bottleneck (GIB) principle, which aims to compress an input graph into a core subgraph that keeps sufficient and compressed information about the input graph \cite{DBLP:conf/iclr/YuXRBHH21}.
However, current GIB methods learn compressed subgraphs by minimizing information loss in predicting the graph labels, which still require the label information.
To solve it, we propose a Subgraph-conditioned Graph Information Bottleneck (S-CGIB) for self-supervised pre-training to compress an input graph into a graph core conditioned on specific significant subgraphs without using label information.
{First}, the graph cores contain important nodes, which could generate robust representations under the S-CGIB principle.
{Second}, toward capturing functional groups without prior knowledge, we suppose that graph cores then could reconstruct input graphs conditioned on specific significant subgraphs across molecules.


To discover the significant subgraphs, as we intentionally ignore using prior knowledge about functional groups, we propose to generate a set of functional group candidates, i.e., ego networks rooted at each node.
The reason is that the graph core of molecules typically consists of central substructures composing important nodes only for generating well-separated representations.
In other words, the compression process ignored unimportant nodes, which can be a part of functional groups in terms of molecular properties.
Then, we propose an attention-based interaction between the graph core and candidates to recognize significant subgraphs as functional groups.
The attention coefficients can highlight significant subgraphs, which benefit from recognizing functional groups across molecules.

\section{Related work}

We now discuss how the existing pre-training strategies can learn the molecular structure and chemical properties in the context of Self-Supervised Learning compared to our method.
Early pre-training strategies focus on learning node representations, which are then aggregated into a single graph-level representation through pooling mechanisms, e.g., min, max, or sum, \cite{DBLP:conf/kdd/HuDWCS20,DBLP:conf/iclr/HuLGZLPL20}.
The node-level strategies can be mainly grouped into node-level structure reconstruction, e.g.,  neighborhood prediction \cite{DBLP:conf/kdd/HuDWCS20,DBLP:conf/aaai/HoangL24}, or node feature reconstruction \cite{DBLP:conf/naacl/DevlinCLT19}.
However, these methods focus on learning to distinguish individual node representations, which do not directly handle the challenge of incorporating node embeddings into a single graph-level representation.
Furthermore, node-level pre-training strategies could be limited to capture the high-order structures, i.e., functional groups.

Another line is contrastive learning, which is generally grouped into two categories: node-level contrastive methods and graph-level contrastive methods \cite{DBLP:conf/iclr/LiuWLLGT22,DBLP:conf/icml/StarkBCTDGL22,DBLP:conf/kdd/QiuCDZYDWT20}.
The node-level contrastive learning strategies seek to generate multiple node views by applying augmentation schemes, e.g., edge perturbation, which modifies the graph connectivity while preserving the node's identity.
For example, GraphCL \cite{DBLP:conf/nips/YouCSCWS20} adopts multiple augmentation schemes to generate multiple views of the input graph and uses contrastive loss to obtain embeddings of these views closer.
The idea of JOAO \cite{DBLP:conf/icml/YouCSW21} is to automatically search schemes to find the most effective augmentations.
In contrast, graph-level methods generate multi-views of an input graph and guarantee similar representations while discriminating them from the other graph-level representations \cite{DBLP:conf/icml/XuWNGT21}.
Such augmentations often change the molecular connectivity and structure, failing to preserve the molecule's properties.


Recent pre-training approaches on molecules emphasize recognizing and learning semantic patterns, such as functional groups, across molecule data \cite{DBLP:conf/nips/ZhangLWLL21,DBLP:conf/nips/LiuSZZKWC23,inae2024motifaware}.
The semantic subgraphs can be discovered under pre-defined rules with the use of prior knowledge or human annotation.
For example, GROVER \cite{DBLP:conf/nips/RongBXX0HH20} extracts 85 frequent functional groups and employs a Self-supervised Learning (SSL) task to predict the presence of the functional groups.
MGSSL \cite{DBLP:conf/nips/ZhangLWLL21} employs depth-first or breadth-first search to discover the semantic subgraphs at multiple scales.
Recently, GraphFP \cite{DBLP:conf/nips/LuongS23} decomposes molecules into bags of functional groups by composing a dictionary via frequent subgraph mining, i.e., common and large frequent subgraphs across molecules.
Such strategies rely on pre-defined vocabulary with discrete counting frequent subgraphs in molecules, making it challenging to build a complete dictionary to cover new or less common functional groups.
In contrast, we automatically discover functional groups via attention-based interaction between the molecular core and a set of functional group candidates.


\section{Problem Descriptions}




We study the problem of pre-trained GNNs on molecules, which recognize graph cores conditioned on significant subgraphs (functional group candidates) under the S-CGIB principle.
Thus, we first present notations and then the definition of S-CGIB, which is a modification of GIB and CGIB.

A molecule can be represented as a graph $G = (V, E)$, where $V= \{v_1, v_2, \cdots, v_N\}$ represents the set of atoms and $E$ denotes the set of bonds. 
$G$ is associated with its adjacency matrix $A$ and feature matrix $X$.
Let $N_k(v)$ be a set of neighboring nodes within a $k$-hop distance from the root node $v$.
The set of functional group candidates in $G$ is defined as: $S = \{G[N_k(v)] \ | \  v \in V \}$, where $G[N_{k}{(v)}]$ is the $k$-hop ego network rooted at node $v$.

Recently, the Information Bottleneck (IB) principle \cite{DBLP:journals/corr/physics-0004057} has been used on graphs, called GIB, to discover a core subgraph from an input graph.
\begin{definition}[GIB]\label{def:GIB} 
The GIB principle was originally introduced by \citet{DBLP:conf/iclr/YuXRBHH21} to recognize a compressed and informative subgraph from an input graph.
Given an input graph $G$ and its label $Y$, the compressed graph, as graph core $G_c$, is discovered as:
\begin{eqnarray}
\label{eq:GIB}
\underset{G_{c }} {\mathop{\min}} - I(Y ;  G_{c })+ \beta I(G ;  G_{c}),
\end{eqnarray}
where $\beta$ is a Lagrange multiplier used to balance the two terms.
The first term encourages $G_{c}$ to be informative to the graph label $Y$, and the second term is the compression term, which minimizes the mutual information of $G$ and $G_{c}$.
\end{definition}

In the context of the presence of side information $T$, several studies have accounted for the conditional information $T$, named CIB \cite{DBLP:conf/nips/ChechikT02,gondek2003conditional}.
We then apply CIB on graphs as Conditional Graph Information Bottleneck (CGIB).
\begin{definition}[CGIB]
\label{def:CGIB}
Given an input graph $G $ and its label $Y$, the graph core $G_c$ is discovered given the side information $T$, as: 
\begin{align}
\label{eq:CGIB}
\underset{G_c} {\mathop{\min}}  - I(Y ; G_c | T )+ \beta I(G ;  G_c) .
\end{align}
The first term quantifies how much information the graph core $G_c$ retains about $Y$, given the side information $T$.
The variable $T$ is supposed to be known as prior knowledge.  
\end{definition}




To employ the CGIB principle under an SSL task without prior knowledge, we propose to minimize the first term of Eq. \ref{eq:CGIB} by replacing the prediction with a reconstruction task.
As the compression process ignored unimportant nodes, which can be a part of functional groups, we suppose they can be side information for a graph reconstruction task.
The key idea is that molecules share universal graph cores, which can reconstruct the molecule structure conditioned on specific significant subgraphs.
Specifically, we suppose that $G$ is formed by combining a graph core $G_{c}$ and a set of functional group candidates $S$, such that $G = G_{c} \cup S$.
Then, the Subgraph-conditioned Graph Information Bottleneck can be defined below:

\begin{definition}[S-CGIB]\label{def:CGIB_SSL}
Given an input graph $G$ and a set of functional group candidates $S$,
we define the S-CGIB principle conditioned on the subgraph $S$ as:
\begin{eqnarray}
\label{eq:S_CGIB}
&\underset{G_c }{\mathop{\min }} -\underbrace{I\left ({G};{{G}_{c}}|{{S}}\right )}_{\text{Conditional Reconstruction}}+\underbrace{\beta I \left({G}; {{G}_{c}} \right )}_{\text{Compression}} , \
\end{eqnarray}
\end{definition}

By conditioning on $S$, the first term encourages the graph core $G_{c}$ to capture sufficient information for reconstructing input graph $G$ (\textit{conditional reconstruction}), while the graph core also needs to be compressed from the input graph (\textit{compression}).
Overall, jointly optimizing the two terms allows ${G}_{c}$ to be compressed and preserve the input graph structure conditioned on ${S}$.
Since $S$ consists of all ego networks rooted at each node, to discover important ego networks, i.e., functional groups, we propose an attention-based strategy detailed in the following Section.

\section{Methodology}

\begin{figure*}[tb]
\centering 
  \includegraphics[width= \linewidth]{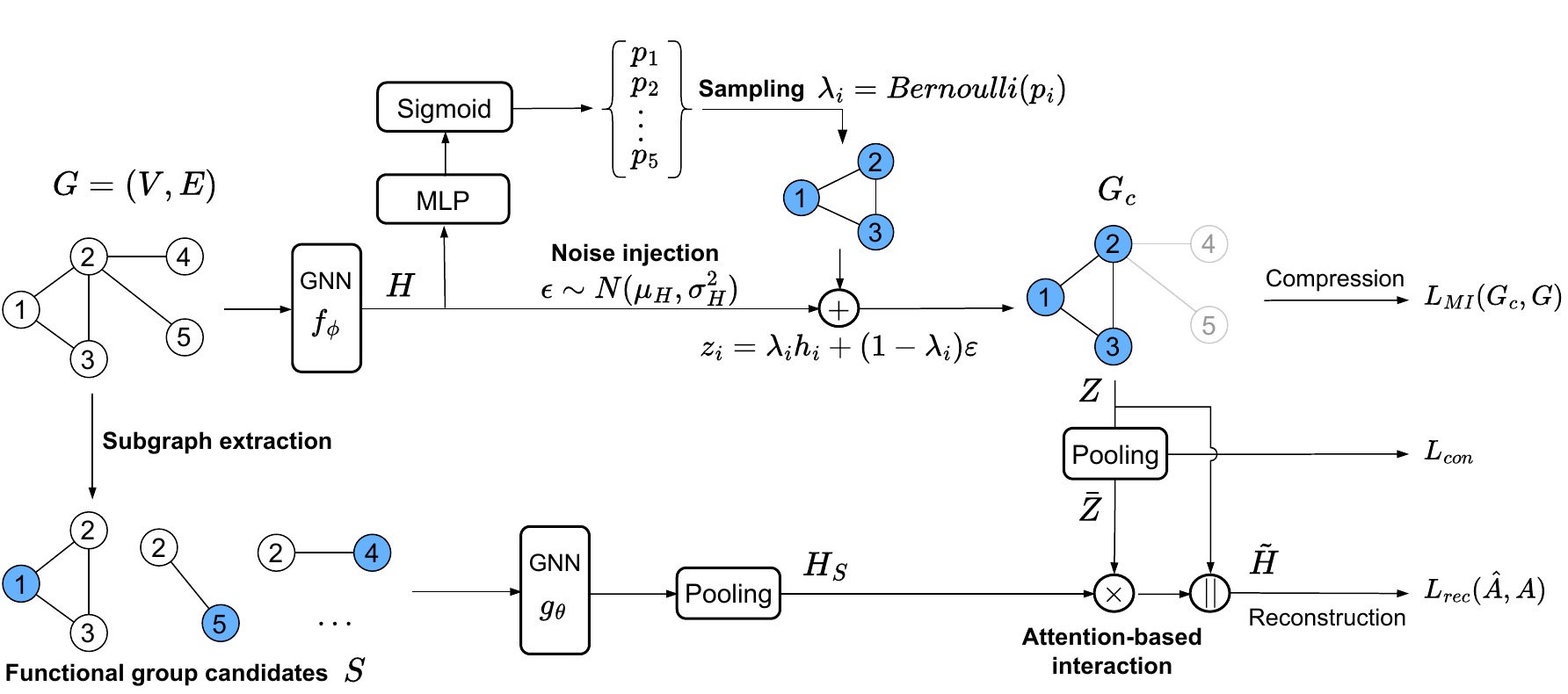}
  \caption{The overall architecture of S-CGIB.}
  \label{fig:model}
\end{figure*}


\subsection{Model Architecture}
\label{sect:model_architecture}

\subsubsection{Graph Compression}
\label{sect:compression}

The overall architecture of S-CGIB is shown in Figure \ref{fig:model}.
Given an input graph $G$ with adjacency matrix $A$ and node feature matrix $X$, we learn node representations in $G$ through a GNN encoder as:
\begin{eqnarray}
&H ={{f}_{\phi}}\left( X,A \right), 
\end{eqnarray}
where $H \in R^{N \times d}$ refers to node embeddings, ${f}_{\phi}(\cdot,\cdot)$ is a GNN encoder, e.g., GIN \cite{DBLP:conf/iclr/XuHLJ19}.
Inspired by recent VGIB principle \cite{DBLP:conf/cvpr/Yu0H22} that injects noise information into an input graph to obtain important nodes, we compress $G$ by injecting noise into $H$ to obtain a graph core $G_c$ with new node representations $Z$, which is a bottleneck to distill the important nodes, thereby generating well-distinguished graph-level representations.
The key idea is that the important nodes will be injected with less noise information compared to unimportant nodes.
Specifically, given the embedding matrix ${H}$, for each node $v_i$, S-CGIB learns a probability $p_i$ with a multi-layer perceptron ($\operatorname{MLP}$) followed by a $\operatorname{Sigmoid}(\cdot)$ function.
We then replace the node representation $h_i$ by $\epsilon$ with probability $p_i$, as:
\begin{eqnarray}
& p_i = \operatorname{Sigmoid} \left ( \operatorname{MLP}({{H}_i}) \right ) ,\\ 
& \varepsilon \sim N\left( {{\mu }_{{H}}},{{\sigma }^2_{H}} \right), \ 
{{\lambda }_{i}} \sim \operatorname{Bernoulli}\left( {p_i} \right), \label{eq:lambda}\\
& {{z}_{i}}={{\lambda }_{i}} {h_i}+(1-{{\lambda }_{i}})\varepsilon  ,
\end{eqnarray}
where $\lambda$ is obtained by sampling from Bernoulli distribution parameterized with the probability $p_i$, 
${{\mu }_{{H}}}$ and ${{\sigma }_{{H}}}$ are the mean and variance of ${H}$,
$\epsilon$ is sampled from ${H}$ based on Gaussian distribution.
Therefore, the information of the input graph is compressed into $Z$ with the probability of $p_i$ by masking unimportant nodes with noisy information.
That is, the compression process ensures that $Z$ focuses on the important nodes (graph core) while discarding irrelevant and unimportant nodes.
To allow the differentiable sampling, we adopt the Gumbel sigmoid method \cite{DBLP:conf/iclr/JangGP17,DBLP:conf/iclr/MaddisonMT17}, i.e.,  ${{\lambda }_{i}}=\operatorname{Sigmoid}( 1/\tau \log [ {{p}_{i}}/(1-{{p}_{i}}) ] )+\log \left[ q/(1-q) \right]$ where $ q \sim \operatorname{Uniform}(0,1)$ and $\tau$ is the temperature parameter.
For the detailed compression optimization process, we refer readers to the Model Optimization Section.


\subsubsection{Subgraph Learning}
\label{sect:subgraph}

The next problem is to extract functional group candidates and encode them to obtain vector representations.
Let $G[N_{k}{(v)}]$ denotes the $k$-hop ego network rooted at the node $v$.
We first apply a GNN encoder on nodes within the $G[N_{k}{(v)}]$.
Then, the subgraph-level representation rooted at node $v$ is obtained via a pooling function $\text{POOL}(\cdot)$.
Formally, the representations of ego networks in the input graph $G$ can be computed as:
\begin{eqnarray}
 h_{v}^{(l+1)}& = & g_{\theta}^{(l)}\left( G \left[ {{N}_{k}}\left( v \right) \right] \right)\text{  , }l=0,\cdots ,L-1 ,\\ 
 h_{v}^{{}}& = & \text{POOL}\left( h_{u}^{(L)} | u \in {{N}_{k}}\left( v \right) \right) , \\ 
{{H}_{S}}& =&[h_{v}^{{}}|v\in V], \ 
\end{eqnarray}
where 
$g_{\theta}^{(l)}(\cdot)$ denotes a GNN encoder, e.g., GIN, for the subgraph $G[N_{k}{(v)}]$, 
$\text{POOL}(\cdot)$ refers to a subgraph pooling, e.g., sum.
Therefore, the local surrounding structures rooted at each node are captured via GNNs.

\subsubsection{Graph Core and Subgraph Interaction}
\label{sec:core_subs_interaction}
Note that each functional group has distinct chemical characteristics that contribute differently to the overall molecule behavior.
To capture significant subgraphs across molecules, we then propose an attention-based interaction between the graph core and subgraph candidates to highlight specific significant subgraphs.
Specifically, we first employ a pooling function to aggregate important node features $Z$ into $\Bar{Z}$, i.e., $\Bar{Z} = \text{POOL}(Z)$, to obtain a single vector representation of the graph core.
We then calculate the normalized attention coefficients for each functional group candidate.
Formally, the coefficient of a subgraph candidate rooted at a node $v_i$ can be computed as:
\begin{align}
& {{\alpha }_{i}}=\frac{\exp\left( \left(\Bar{Z} || H_{S_i} \right) W_{a}^{T}\right)}{\sum\nolimits_{k=1}^{N}{ \exp\left( \left(\Bar{Z} || H_{S_k} \right)W_{a}^{T}\right)}}, 
\end{align}
where $W_a \in R^{1 \times 2d}$ refers to learnable projection and $H_{S_i}$ is the representations of the ego network rooted at node $v_i$.
Thus, the interaction could capture the correlation between the graph core and functional group candidates.
The final representations are then concatenated along with the coefficients as:
\begin{align}
&\tilde{H}_{i}= Z_{i} || \left( {{\alpha }_{i}} H_{S_i} \right). 
\end{align}

In a nutshell, given an input graph $G$, our model jointly learns the graph core and the significant subgraphs under the S-CGIB principle.
We first inject noisy information into $H$ to obtain the graph core $Z$ under the compression process.
Meanwhile, we capture the significant specific subgraphs from the subgraph candidates (ego networks rooted at each node) through the attention-based interaction between the graph core and the ego networks under the graph reconstruction.



\subsection{Model Optimization}
\label{sect:optimization}

To train the model while optimizing the graph core conditioned on $S$, we optimize the objective function:
\begin{eqnarray}
\label{eq:cgib_opt}
&G^{*}_c =\underset{{{G}_{c}}}{\mathop{\arg \min }}\,-I({G};{{G}_{c}}|{{S}})+\beta I({G}; {{G}_{c}}) , \
\end{eqnarray}
where each term denotes the conditional graph reconstruction and compression, respectively.
Then, we present upper bounds for each term to guide the optimization.

\subsubsection{Minimizing  $-I({G};{{G}_{c}}|{{S}})$}
\label{sect:opt_1term}

The first term of  Eq. \ref{eq:cgib_opt} denotes a reconstruction of the input graph $G$, given $G_c$ conditioned on $S$.
Thus, we utilize the chain rule for mutual information on the Conditional Graph Reconstruction term as follows:
\begin{eqnarray}
\label{eq:min11}
& \min -I({G};{{G}_{c}}|{{S}})= \min -I({G};{{G}_{c}},{{S}})+I({G};{{S}}).
\end{eqnarray}%

We observed that including the second term, i.e., $I({G};{{S}})$, into our objectives severely degrades the overall model performance (Appendix D.1).
That is, the model performs worse as we push the input graph $G$ and its functional group candidates $S$ far apart.
Therefore, we only minimize the first term of Eq. \ref{eq:min11}.

The first term of Eq. \ref{eq:min11} can be bounded as follows:
\begin{eqnarray}
\label{eq:min12}
& -I({G};{{G}_{c}},{{S}})\le {\mathbb{E}_{{G};{{G}_{c}},{{S}}}} [ -\log {{p}_{\varsigma }}( {G}|{{G}_{c}},{{S}} ) ] ,   \
\end{eqnarray}%
where ${{p}_{\varsigma }}\left( {G}|{{G}_{c}},{{S}} \right)$ is a variational approximation of $p({G}|{{G}_{c}},{{S}})$, which outputs the input graph $G$ (see Appendix A.1).
Thus, we model ${{p}_{\varsigma  }} \left( {G}|{{G}_{c}},{{S}} \right)$ as a graph structure reconstruction parametrized by $\varsigma $, which outputs the graph $G$ based on the $G_c$ and ${S}$.
Therefore, we can minimize the upper bound of $-I({G};{{G}_{c}},{{S}})$ by minimizing the graph reconstruction loss $L_{rec}({G};{{G}_{c}},{{S}})$, which can be modeled as graph structure recovery.
Specifically, given the output representation $\tilde{H}$, to minimize the graph structure loss, we first capture the similarity between any two nodes $v_i$ and $v_j$ in $\tilde{H}$ by employing cosine similarity, i.e., ${{\hat{A}}_{i,j}}=\frac{{{{\tilde{H}}}^{\top}}_{i}{{{\tilde{H}}}_{j}}}{\| {{{\tilde{H}}}_{i}} \|\| {{{\tilde{H}}}_{j}} \|}$, \cite{zhang2020graph}. 
The reconstruction loss then can be defined as follows:
\begin{eqnarray}
& L_{rec}=\frac{1}{{{\left| V \right|}^{2}}}\| A- \hat{A} \|_{F}^{2} \ ,   
\end{eqnarray}
where $A$ refers to the original adjacency matrix of $G$ and $||\cdot||_F$ is the Frobenius norm.

\subsubsection{Minimizing $I(G ; G_c)$}
\label{sect:compression}

To minimize the second term of Eq. \ref{eq:cgib_opt}, we employ the sufficient encoder assumption that the latent representation $Z$ is lossless in the encoding process, i.e., $I( Z|H ) \approx I({{G}_{c}}|{G})$.
To optimize the graph core $G_c$, we can employ a variational upper bound that is tractable and can be minimized during training \cite{DBLP:conf/cvpr/Yu0H22}.
Formally, given the mutual information $I(G ; G_c)$, the variational upper bound of $I(G ; G_c)$ is:
\begin{equation}
\label{eq:mi1}
\resizebox{0.415 \textwidth}{!}{${{L}_{MI}}\left( G, G_c \right)\le {{\mathbb{E}} }\left( -\frac{1}{2}\log {{P} }+\frac{1}{2N}\log {{P} }+\frac{1}{2N}\log Q^{2} \right)$} ,
\end{equation}%
where ${{P} }=\sum\nolimits_{j=1}^{N}{{{\left( 1-{{\lambda }_{j}} \right)}^{2}}}$,
${{Q} }=\frac{\sum\nolimits_{j=1}^{N}{{{\lambda }_{j}}{{\left( {{H}_{j}}-{{\mu }_{H}} \right)} }}}{{{\sigma }_{H}}} $,
and $\lambda$ is computed from Eq. \ref{eq:lambda}.
For the proof of the upper bound, we refer readers to Appendix A.2.



\subsubsection{Contrastive Learning}
We note that minimizing the upper bound of $I(G,  G_c)$ from Eq. \ref{eq:mi1} could lead to over-compression with sufficient information on $S$.
That is, the graph core $G_c$ can be too distinguished from its input graph $G$ compared to other graphs.
We argue that the ideal core should at least satisfy the high mutual information with its input graph compared to others during the compression process, as $I\left( G_{c}, G \right) \ge  I\left( G_c, \setminus G \right)$,
where $\setminus G$ refers to remaining graphs, excluding $G$. 
Thus, we propose to use a contrastive learning-based method to maximize the agreement between the graph core and its input graph.
Specifically, the contrastive objective is computed as:
\begin{equation}
L_{con} = - \frac{1}{B}\sum\limits_{i=1}^{B}{\log \frac{\exp \left( s\left( \Bar{Z}^{i}, \Bar{H}^{i} \right) \right)}{\sum\nolimits_{j=1, j\ne i}^{B}{\exp \left( s \left( \Bar{Z}^{i}, \Bar{H}^{j} \right) \right)}}},  \
\end{equation}
where $B$ denotes the number of graphs in a mini-batch, 
$s(\cdot, \cdot)$ is the cosine similarity between graph core and input graph, $\Bar{Z}= \text{POOL}(Z)$, and $\Bar{H}= \text{POOL}(H)$.

The overall loss for the pre-training task is as follows:
\begin{eqnarray}
L_{total}=   {L}_{con} +  {{L}_{rec}} + \beta {L}_{MI}, \
\end{eqnarray}
where $ \beta$ is a hyperparameter to balance the compression and structure preservation trade-off.

\subsubsection{Domain Adaptation}

We note that our pre-trained model can learn significant subgraphs only on the domains of pre-training datasets.
However, generalizing toward varied downstream tasks can be challenging due to the node attribute distinctions in specific downstream datasets.
That is, while pre-training primarily focuses on structural molecular features, node feature reconstruction is also essential and helps our model adapt to learn node feature characteristics in downstream datasets.
Therefore, we employ an unsupervised domain adaptation after pre-training, which acts as a domain-oriented generalization for downstream datasets.
Then, we utilize a loss function to reconstruct node feature information for each graph as:
\begin{eqnarray}
{{L}_{att}}=\frac{1}{|V|}\sum\limits_{{{v}_{i}}\in V}{{{\left\| {{x}_{i}}-{\hat {x}_{i}} \right\|}_{2}}}, 
\end{eqnarray}
where $x_i$ is the initial feature of node $v_i$ and ${\hat {x}_{i}} = \text{MLP}(\tilde{H_i})$.

\section{Evaluation}



\subsection{Experimental Settings}

\subsubsection{Datasets}

We conducted experiments across various molecular domains to evaluate the effectiveness of our proposed model, including  
Biophysics (mol-HIV, mol-PCBA, and BACE \cite{wu2018moleculenet}),
Physiology (BBBP, Tox21, ToxCast, SIDER, ClinTox, and MUV),
Physical Chemistry (ESOL, FreeSolv, and Lipophilicity) \cite{wu2018moleculenet},
Bioinformatics (Mutagenicity, NCI1, and NCI109 \cite{Morris_2020}),
Quantum Mechanics (PCQ4Mv2 and QM9 \cite{DBLP:conf/nips/HuFRNDL21}).
For pre-training datasets, we considered 300k unlabeled molecules sampled from three datasets, i.e., PCQ4Mv2, QM9, and mol-PCBA.
The remaining datasets are used as fine-tuning datasets.
Furthermore, we also evaluated the model's performance on large molecular graphs using two peptide molecules: peptide-func and peptide-struct \cite{DBLP:conf/nips/DwivediRGPWLB22}.
For the model explainability, we utilized four datasets with the availability of ground truth, i.e., Mutagenicity, Benzene, Alkane Carbonyl, and Fluoride Carbonyl \cite{agarwal2023evaluating}.
We randomly split the datasets into training/validation/test sets with a ratio of 6:2:2.
The datasets are given in Appendix B.





\subsubsection{Baselines and Implementation Details}
We considered three groups of baselines.
(i) The node-level pretraining methods are 
ContextPred and AttrMasking \cite{DBLP:conf/iclr/HuLGZLPL20}, 
and EdgePred \cite{DBLP:journals/corr/HamiltonYL17}.
(ii) The contrastive learning methods are 
Infomax \cite{velickovic2019deep}, 
JOAO and JOAOv2 \cite{DBLP:conf/icml/YouCSW21}, 
GraphCL \cite{DBLP:conf/nips/YouCSCWS20}, 
and GraphLoG \cite{DBLP:conf/icml/XuWNGT21}.
(iii) The subgraph-based methods are 
MICRO-Graph \cite{DBLP:conf/aaai/Subramonian21}, 
MGSSL \cite{DBLP:conf/nips/ZhangLWLL21}, 
GraphFP \cite{DBLP:conf/nips/LuongS23}, 
GROVE \cite{DBLP:conf/nips/RongBXX0HH20}, 
SimSGT \cite{DBLP:conf/nips/LiuSZZKWC23}, 
and MoAMa \cite{inae2024motifaware}.
We adopted a 5-layer GIN \cite{DBLP:conf/iclr/XuHLJ19} as graph encoders.
The hyperparameters and experimental setups are given in Appendix C.
The open-source implementation of S-CGIB is available for reproducibility\footnote{ https://github.com/NSLab-CUK/S-CGIB}.



\subsection{Performance Analysis}

\begin{table*}[t]
\centering
\setlength{\tabcolsep}{4 pt}
\fontsize{9 pt}{9 pt}\selectfont
\caption{A performance comparison on graph classification tasks in terms of ROC-AUC. (D.A.: Domain Adaptation).
}
\begin{tabular}{l cc cc cc c c }
\toprule
Methods
&BBBP
&Tox21
&ToxCast 
&SIDER
&ClinTox
&MUV
&HIV 
&BACE 
\\ 
\midrule
ContextPred 
&69.10$\pm$0.29 
&73.26$\pm$0.59 
&63.28$\pm$0.68 
&61.83$\pm$0.60
&55.63$\pm$1.35 
&71.43$\pm$0.79 
&72.04$\pm$0.48 
&78.39$\pm$0.58 

\\
AttrMasking 
&67.12$\pm$0.45 
&73.37$\pm$0.55 
&61.66$\pm$1.20 
&61.21$\pm$0.65 
&60.11$\pm$1.19 
&67.93$\pm$0.56 
&72.71$\pm$0.70 
&75.95$\pm$0.50 

\\
EdgePred 
&64.73$\pm$1.10 
&70.32$\pm$1.62 
&60.04$\pm$0.81 
&60.18$\pm$0.76 
&61.62$\pm$1.25 
&70.81$\pm$1.58 
&70.55$\pm$1.68 
&74.29$\pm$1.37 

\\
\midrule
Infomax 
&68.39$\pm$0.64 
&72.66$\pm$0.16 
&62.76$\pm$0.54 
&59.02$\pm$0.56 
&58.62$\pm$0.83 
&72.14$\pm$1.25 
&73.55$\pm$0.47 
&77.80$\pm$0.46 

\\
JOAO 
&71.63$\pm$1.11
&73.67$\pm$1.06
&63.30$\pm$0.27
&{63.55$\pm$0.81}
&77.02$\pm$1.64
&69.81$\pm$0.26
&77.55$\pm$1.94 
&74.94$\pm$1.35 

\\
JOAOv2 
&71.98$\pm$0.18
&73.95$\pm$1.88
&63.12$\pm$1.90 
&59.88$\pm$1.72
&65.22$\pm$0.75
&68.50$\pm$1.58
&77.13$\pm$1.51 
&74.38$\pm$1.71 

\\
GraphCL 
&68.39$\pm$0.64 
&73.26$\pm$0.59 
&62.76$\pm$0.54 
&61.83$\pm$0.60 
&61.62$\pm$1.25 
&72.14$\pm$1.25 
&73.55$\pm$0.47 
&77.80$\pm$0.46 

\\
GraphLoG 
&66.75$\pm$0.32
&71.64$\pm$0.49 
&61.53$\pm$0.35 
&59.09$\pm$0.53 
&53.76$\pm$0.95 
&72.52$\pm$2.02 
&73.76$\pm$0.29 
&76.60$\pm$1.04 

\\
\midrule
GraphFP 
&72.05$\pm$1.17
&{77.35$\pm$1.40}
&{69.15$\pm$1.92} 
&\textbf{65.93$\pm$3.09}
&76.80$\pm$1.83
&71.82$\pm$1.33
&75.71$\pm$1.39 
&80.28$\pm$3.06 

\\
MICRO-Graph
&67.21$\pm$1.85
&71.79$\pm$1.70
&60.80$\pm$1.15 
&60.34$\pm$0.96
&{77.56$\pm$1.56}
&70.46$\pm$1.62
&76.73$\pm$1.07 
&63.57$\pm$1.55 

\\ 
MGSSL
&79.52$\pm$1.98
&74.82$\pm$1.60
&63.86$\pm$1.57 
&57.46$\pm$1.45
&75.84$\pm$1.82
&{73.44$\pm$3.47}
&77.45$\pm$2.94 
&{82.03$\pm$3.79}

\\ 
GROVE 
&{87.15$\pm$0.06}
&68.59$\pm$0.24
&64.45$\pm$0.14 
&57.53$\pm$0.23
&72.53$\pm$0.14
&67.67$\pm$0.12
&75.04$\pm$0.13 
&81.13$\pm$0.14 

\\ 
SimSGT 
&71.51$\pm$1.75 
&76.23$\pm$1.27 
&65.83$\pm$0.79 
&59.74$\pm$1.32 
&74.11$\pm$1.05 
&{72.79$\pm$1.52}
&{78.13$\pm$1.07} 
&79.75$\pm$1.28 

\\
MoAMa 
&{85.89$\pm$0.61}
&{78.29$\pm$0.55}
&{68.01$\pm$1.07}
&62.69$\pm$0.37
&{77.11$\pm$1.67}
&72.41$\pm$1.76
&{78.11$\pm$0.64}
&{81.32$\pm$1.06}

\\ 
\midrule
S-CGIB w/o D.A.
&86.71$\pm$0.74	
&79.52$\pm$0.71	
&68.93$\pm$0.45
&62.76$\pm$1.83	
&74.69$\pm$1.28
&74.12$\pm$1.85%
&77.41$\pm$1.63
&86.51$\pm$1.49	
\\ 

S-CGIB 
&\textbf{88.75$\pm$0.49}
&\textbf{80.94$\pm$0.17}
&\textbf{70.95$\pm$0.27} 
&{64.03$\pm$1.04} 
&\textbf{78.58$\pm$2.01}
&\textbf{77.71$\pm$1.19}  
&\textbf{78.33$\pm$1.34}
&\textbf{86.46$\pm$0.81}

\\
\bottomrule
\end{tabular}
\label{tab:acc1}
\end{table*}

\begin{table}[tb]
\caption{A performance comparison on graph classification tasks in terms of accuracy.}
\fontsize{9 pt}{9 pt}\selectfont
\setlength{\tabcolsep}{ 1 mm }
\centering
\begin{tabular}{l ccc  }
\toprule
Methods
&Mutagenicity
&NCI1
&NCI109 
\\
\midrule
ContextPred 
&57.95$\pm$1.42
&49.47$\pm$1.12
&50.32$\pm$1.05 
\\
AttrMasking 
&58.03$\pm$1.16
&49.51$\pm$1.21
&46.25$\pm$1.73 
\\
EdgePred 
&48.58$\pm$1.02
&49.88$\pm$1.12
&49.19$\pm$1.36
\\
\midrule
Infomax 
&56.64$\pm$1.77
&49.55$\pm$1.14
&53.03$\pm$1.48 
\\
JOAO 
&62.33$\pm$1.13
&49.03$\pm$1.21
&58.23$\pm$1.49 
\\
JOAOv2 
&63.36$\pm$1.74
&50.73$\pm$1.74
&53.75$\pm$1.32 
\\
GraphCL 
&66.32$\pm$3.62
&49.11$\pm$1.31
&56.62$\pm$1.59 
\\
GraphLoG 
&66.47$\pm$1.47
&60.94$\pm$1.93
&57.52$\pm$1.61
\\

\midrule
GraphFP
&68.43$\pm$1.32
&53.77$\pm$1.13
&58.14$\pm$1.45 
\\
MICRO-Graph
&{80.64$\pm$1.28}
&74.45$\pm$1.51
&{76.15$\pm$3.53}
\\
MGSSL 
&66.47$\pm$1.47
&60.94$\pm$1.93
&57.52$\pm$1.61
\\
GROVE
&{80.49$\pm$0.94}
&{75.79$\pm$0.91}
&{76.01$\pm$0.73}
\\
SimSGT
&68.27$\pm$0.53
&56.93$\pm$0.43
&60.48$\pm$0.31
\\
MoAMa 
&80.37$\pm$0.87
&{78.59$\pm$0.81}
&{76.82$\pm$1.05} 
\\
\midrule
S-CGIB w/o D.A.
&80.26$\pm$0.71
&78.51$\pm$1.06
&77.08$\pm$1.55



\\

S-CGIB (Ours)
&\textbf{81.12$\pm$0.90} 
&\textbf{79.75$\pm$0.82}
&\textbf{77.54$\pm$1.51}
\\ 
\bottomrule
\end{tabular}
\label{tab:classification_task_acc}
\end{table}

\begin{table}[tb]
\caption{A performance comparison on regression tasks in terms of RMSE.}
\fontsize{9 pt}{9 pt}\selectfont
\setlength{\tabcolsep}{ 1 mm }
\centering
\begin{tabular}{l ccc}
\toprule
Methods
&FreeSolv 
&ESOL
&Lipophilicity
\\
\midrule
ContextPred 
&3.195$\pm$0.058 
&2.190$\pm$0.026
&1.053$\pm$0.048
\\
AttrMasking 
&4.023$\pm$0.039 
&2.954$\pm$0.087 
&0.982$\pm$0.052
\\
EdgePred 
&3.192$\pm$0.023 
&2.368$\pm$0.070
&1.085$\pm$0.061
\\
\midrule
Infomax 
&3.033$\pm$0.026 
&2.953$\pm$0.049
&0.970$\pm$0.023
\\
JOAO 
&3.282$\pm$0.002 
&1.978$\pm$0.029
&1.093$\pm$0.097
\\
JOAOv2 
& 3.842$\pm$0.012
& 2.144$\pm$0.009
&1.116$\pm$0.024
\\
GraphCL 
&3.166$\pm$0.027
&1.390$\pm$0.363 
&1.014$\pm$0.018
\\
GraphLoG 
&2.335$\pm$0.052 
&1.542$\pm$0.026
&0.932$\pm$0.052
\\

\midrule
GraphFP
&2.528$\pm$0.016 
&2.136$\pm$0.096
&1.371$\pm$0.058
\\
MICRO-Graph
&{1.865$\pm$0.061} 
&{0.842$\pm$0.055}
&0.851$\pm$0.073
\\
MGSSL 
&2.940$\pm$0.051 
&2.936$\pm$0.071
&1.106$\pm$0.077
\\
GROVE
&2.712$\pm$0.327 
&1.237$\pm$0.403
&0.823$\pm$0.027
\\
SimSGT
&{1.953$\pm$0.038}
&{0.932$\pm$0.026}
&0.771$\pm$0.041
\\
MoAMa 
&2.072$\pm$0.053
&1.125$\pm$0.029
&1.085$\pm$0.024
\\
\midrule
S-CGIB w/o D.A.

&1.832$\pm$0.095
&0.894$\pm$0.052
&0.803$\pm$0.067


\\

S-CGIB (Ours)
&\textbf{1.648$\pm$0.074} 
&\textbf{0.816$\pm$0.019}
&\textbf{0.762$\pm$0.042}
\\ 
\bottomrule
\end{tabular}
\label{tab:regression_task}
\end{table}

Tables \ref{tab:acc1} and \ref{tab:classification_task_acc} present the performance of our proposed model and the baselines on the graph classification task. 
We observed that:
{(i)} S-CGIB consistently outperformed other baselines, obtaining the best performance in 10 out of 11 downstream datasets in the graph classification tasks.
For the graph regression tasks, we can observe that S-CGIB consistently outperformed other methods on three regression benchmarks, as shown in Table \ref{tab:regression_task}.
We attribute the superior performance of S-CGIB to its ability to generate well-distinguished representations and effectively capture significant substructures, i.e., functional groups.
For example, in the BBBP dataset, the task is to predict the barrier permeability, where molecular structures with different sizes, skeletal ring structures, and functional groups will decide the penetrating properties.
Capturing such structural information to generate graph-level molecular representations can enhance the prediction of molecular properties.
{(ii)} Subgraph-level strategies, e.g., GraphFP and MGSSL, outperformed node-level and contrastive learning methods.
This implies that subgraph-level methods could capture well global molecular structure and functional groups, which benefits molecular property prediction in downstream tasks.
In contrast, S-CGIB not only learns well-separated representations but also automatically captures functional groups.


\subsubsection{Performance on Large Molecular Graphs.}
To validate the model's ability on large molecules, we conducted experiments on two large molecular graphs, i.e., Peptides-func and Peptides-struct, as shown in Table \ref{tab:LRGB}.
The results demonstrated that S-CGIB outperformed other baselines, including subgraph-based strategies.
For example, on the Peptides-func dataset, our proposed model gained a 12.3\% improvement compared to the GraphFP method.
It indicates that S-CGIB could capture long-range dependencies by generating well-separated representations and then capturing significant subgraphs, thanks to the attention-based interaction between graph core and significant subgraphs.
The results verified the effectiveness of our strategy for capturing well-separated representations and significant subgraphs.

\begin{table*}[tb]
\caption{An interpretability comparison on functional group detection tasks in terms of $Fidelity-$/$+$.}
\centering
\setlength{\tabcolsep}{ 1 mm }
\fontsize{9 pt}{9 pt} \selectfont
\begin{tabular}{l cc cc cc cc}
\toprule
Methods
&\multicolumn{2}{c}{Mutagenicity}
&\multicolumn{2}{c}{BENZENE} 
&\multicolumn{2}{c}{Alkane Carbonyl} 
&\multicolumn{2}{c}{Fluoride Carbonyl} 
\\
 \cmidrule(lr){2-3}\cmidrule(lr){4-5}
\cmidrule(lr){6-7}\cmidrule(lr){8-9}
 
&$Fidelity-\downarrow$
&$Fidelity+\uparrow$
&$Fidelity-\downarrow$
&$Fidelity+\uparrow$
&$Fidelity-\downarrow$
&$Fidelity+\uparrow$
&$Fidelity-\downarrow$
&$Fidelity+\uparrow$
\\
\midrule
ContextPred
&{0.061$\pm$0.002}
&0.223$\pm$0.004

&0.419$\pm$0.008
&0.483$\pm$0.005

&0.261$\pm$0.001
&0.293$\pm$0.007

&0.363$\pm$0.018
&0.413$\pm$0.024
\\

AttrMasking
&0.078$\pm$0.005
&0.230$\pm$0.004

&0.448$\pm$0.002
&0.543$\pm$0.016

&0.260$\pm$0.011
&0.310$\pm$0.009

&0.276$\pm$0.007
&0.384$\pm$0.005
\\
EdgePred
&0.081$\pm$0.003
&0.451$\pm$0.013

&0.386$\pm$0.068
&0.457$\pm$0.061

&0.581$\pm$0.074
&{0.603$\pm$0.069}

&0.342$\pm$0.073
&0.389$\pm$0.072
\\

\midrule
Infomax 
&{0.064$\pm$0.004}
&0.240$\pm$0.008

&0.363$\pm$0.041
&0.410$\pm$0.082

&0.353$\pm$0.021
&0.376$\pm$0.054

&0.331$\pm$0.032
&0.453$\pm$0.012
\\
JOAO
&0.103$\pm$0.004
&0.424$\pm$0.013

&{0.047$\pm$0.005}
&0.481$\pm$0.007

&0.263$\pm$0.005
&0.568$\pm$0.008

&0.183$\pm$0.007
&0.295$\pm$0.004
\\
JOAOv2
&0.152$\pm$0.005
&0.431$\pm$0.008

&0.062$\pm$0.006
&0.481$\pm$0.009

&0.387$\pm$0.004
&0.586$\pm$0.007

&{0.184$\pm$0.007}
&0.207$\pm$0.008
\\
GraphCL
&0.283$\pm$0.008
&0.476$\pm$0.002

&0.120$\pm$0.005
&0.469$\pm$0.008

&0.430$\pm$0.027
&0.578$\pm$0.016

&0.284$\pm$0.002
&0.570$\pm$0.001
\\

GraphLoG
&0.117$\pm$0.001 
&0.439$\pm$0.003 

&0.137$\pm$0.000 
&0.459$\pm$0.007 

&0.355$\pm$0.002 
&0.695$\pm$0.007

&0.358$\pm$0.004 
&0.475$\pm$0.006
\\

\midrule

GraphFP
&0.213$\pm$0.025 
&0.603$\pm$0.003

&0.093$\pm$0.006 
&{0.494$\pm$0.012}

&0.276$\pm$0.022 
&0.529$\pm$0.005

&0.263$\pm$0.026 
&0.595$\pm$0.034
\\
MICRO-Graph
&0.235$\pm$0.008 
&0.466$\pm$0.016 

&0.140$\pm$0.012 
&0.483$\pm$0.003 

&{0.155$\pm$0.037}
&0.524$\pm$0.015

&0.281$\pm$0.007 
&{0.595$\pm$0.002}
\\
MGSSL 

&0.150$\pm$0.005
&0.489$\pm$0.006

&\textbf{0.019$\pm$0.001}
&0.480$\pm$0.006

&\textbf{0.049$\pm$0.001}
&0.205$\pm$0.002

&0.231$\pm$0.001
&0.550$\pm$0.006

\\
GROVE
&0.218$\pm$0.005
&0.522$\pm$0.014

&0.064$\pm$0.029 
&0.489$\pm$0.010 

&0.183$\pm$0.041 
&0.518$\pm$0.007

&0.321$\pm$0.067 
&{0.628$\pm$0.038}
\\
MoAMa 
&0.228$\pm$0.020 
&{0.585$\pm$0.012}

&0.076$\pm$0.006 
&0.483$\pm$0.002 

&0.274$\pm$0.021
&0.514$\pm$0.005

&0.220$\pm$0.007 
&0.451$\pm$0.008
\\
\midrule
S-CGIB (Ours)
&\textbf{0.008$\pm$0.001}
&\textbf{0.638$\pm$0.003}	

&{0.049$\pm$0.001}
&\textbf{0.720$\pm$0.003}	

&{0.134$\pm$0.001}
&\textbf{0.727$\pm$0.003}	

&\textbf{0.133$\pm$0.002}
&\textbf{0.672$\pm$0.004}	
\\
\bottomrule
\end{tabular}
\label{tab:XGNNs}
\end{table*}

\begin{table}[tb]
\caption{A performance comparison on the two large molecular graph datasets.}
\fontsize{9 pt}{9.7  pt}\selectfont
\setlength{\tabcolsep}{ 6 pt}
\centering
\begin{tabular}{l cc  }
\toprule
Methods
& Peptides-func
& Peptides-struct 
\\
& (AP$\uparrow$) 
& (MAE$\downarrow$) 
\\
\midrule
ContextPred 
&0.311$\pm$0.013
&0.587$\pm$0.001
\\
AttrMasking 
&0.318$\pm$0.002
&0.580$\pm$0.002
\\
EdgePred 
&0.310$\pm$0.012
&0.546$\pm$0.001
\\
\midrule
Infomax 
&0.335$\pm$0.013
&0.574$\pm$0.001
\\
JOAO 
&0.386$\pm$0.009
&0.463$\pm$0.008
\\
JOAOv2 
&0.398$\pm$0.009
&0.541$\pm$0.008
\\
GraphCL 
&0.380$\pm$0.002
&0.973$\pm$0.014
\\
GraphLoG 
&0.313$\pm$0.034
&0.419$\pm$0.006
\\

\midrule
GraphFP
&{0.618$\pm$0.014}
&{0.327$\pm$0.026}
\\
MICRO-Graph
&0.505$\pm$0.014
&{0.332$\pm$0.002}
\\
MGSSL 
&0.541$\pm$0.006
&{0.322$\pm$0.008}
\\
GROVE
&0.587$\pm$0.023
&0.376$\pm$0.005
\\
SimSGT
&{0.612$\pm$0.005}
&0.358$\pm$0.003
\\
MoAMa 
&0.584$\pm$0.019
&0.365$\pm$0.005
\\
\midrule
S-CGIB w/o D.A.


&0.658$\pm$0.013
&0.306$\pm$0.007

\\
S-CGIB (Ours)
&\textbf{0.694$\pm$0.002}
&\textbf{0.269$\pm$0.004}
\\ 
\bottomrule
\end{tabular}
\label{tab:LRGB}
\end{table}

\begin{figure}[t]
\centering
  \includegraphics[width= \linewidth]{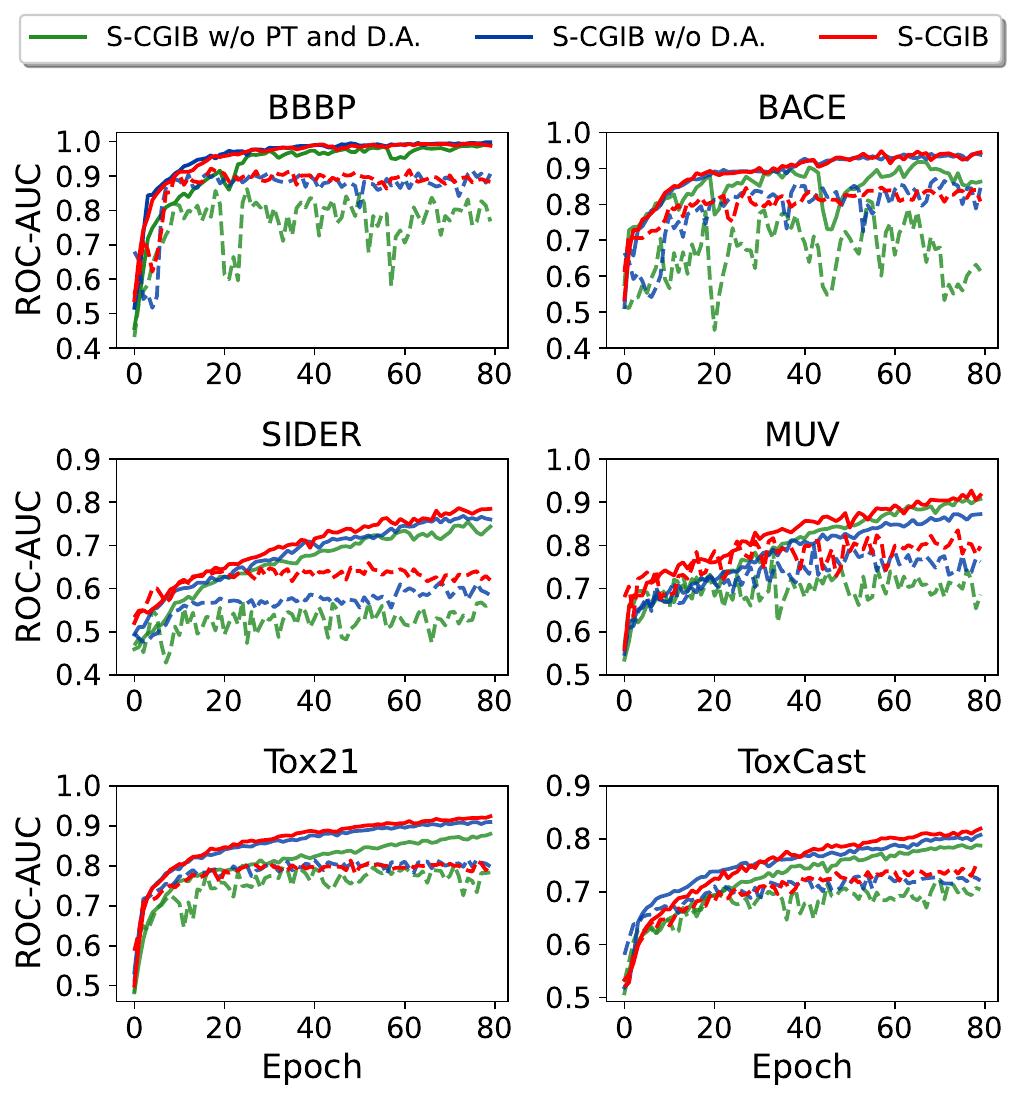}
  \caption{An efficiency analysis for variants of S-CGIB.
  The solid lines are training curves, and the dashed lines are validation curves (PT: Pre-training, D.A.: Domain Adaptation).}
  \label{fig:efficiency}
\end{figure}

\subsubsection{Efficiency Analysis.}

Beyond performance improvement, we also validated the impacts of the pre-trained model's convergence and generalization.
Figure \ref{fig:efficiency} shows that S-CGIB almost converged faster than that without pre-training and domain adaptation.
This is because S-CGIB has already captured the important patterns from the pre-training dataset well.
Besides, our pre-trained model is considered a one-time effort, which can significantly reduce the training and validation time on specific downstream datasets.
Moreover, S-CGIB with pre-training and domain adaptation typically exhibits stable training and validation curves.


\begin{figure*}[t]
\centering 
  \includegraphics[width= \linewidth]{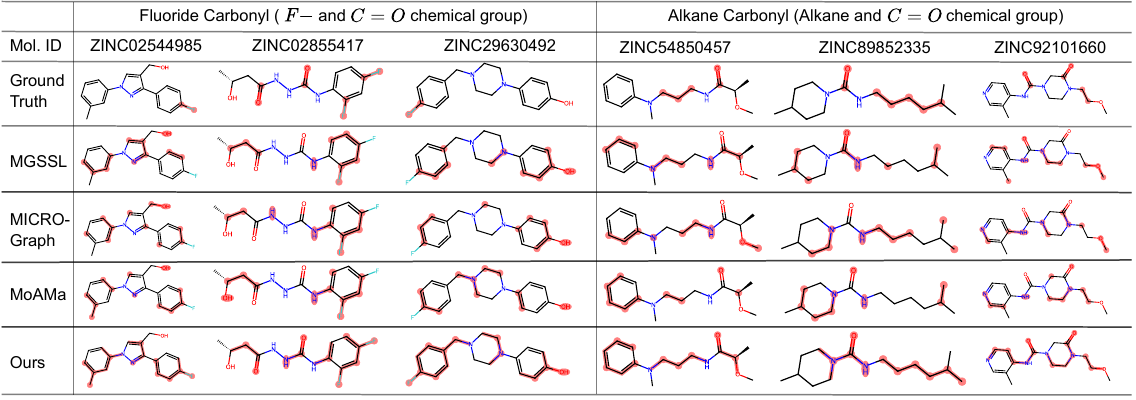}
  \caption{Visualizations of model interpretability in functional group detection tasks.
  }
  \label{fig:Qualitative}
\end{figure*}



\subsubsection{Analysis on Distinguishability of Representations}

To validate the model’s ability to generate well-distinguished representations, we further conducted experiments to validate the well-distinguished representations between different molecular structures.
We utilized Jensen-Shannon Divergence (JSD) measurement to validate the distinction between learned embeddings, as shown in Table \ref{tab:well_distinction}.
We adopted three datasets: BENZENE, Alkane Carbonyl, and Fluoride Carbonyl datasets, with ground-truth explanations, whose classes are labeled based on structures.
We observed that our proposed model learned well-distinguished and robust representations. 
This robustness enhances the discriminability of molecular representations, confirming the effectiveness of our proposed model in capturing different molecular structures across molecules.

\begin{table}[t]
\caption{
A distinguishability comparison in terms of JSD.
}
\centering
\setlength{\tabcolsep}{1 mm}
\fontsize{9 pt}{10 pt}\selectfont
\begin{tabular}{l ccc}
\toprule
Methods
&BENZENE
&Alkane Carbonyl
&Fluoride Carbonyl
\\ 
\midrule
MGSSL
&0.201$\pm$0.008
&0.253$\pm$0.012
&0.218$\pm$0.011
\\
SimSGT	
&0.173$\pm$0.010	
&0.124$\pm$0.014	
&0.137$\pm$0.007
\\
MoAMa	
&0.215$\pm$0.016	
&0.209$\pm$0.013
&0.153$\pm$0.015
\\ 
\midrule
S-CGIB 
&\textbf{0.381$\pm$0.009}	
&\textbf{0.362$\pm$0.005}	
&\textbf{0.295$\pm$0.006}
\\ 
\bottomrule
\end{tabular}

\label{tab:well_distinction}
\end{table}

\begin{table}[t]
\caption{
An ablation analysis on graph core ($G_c$) and attention coefficients ($\alpha$).
}
\centering
\setlength{\tabcolsep}{ 6 pt}
\fontsize{9 pt}{10 pt}\selectfont
\begin{tabular}{cc ccc }
\toprule
$G_{c}$
&$\alpha$
&BBBP
&Tox21
&ToxCast
\\ 
\midrule
$-$ 
&$-$ 
&81.15$\pm$0.75
&78.68$\pm$0.63 
&68.39$\pm$0.48 
\\

$\checkmark$
& $-$ 
&{86.26$\pm$0.63} 
&{79.48$\pm$0.49} 

&{69.04$\pm$0.38}


\\
$-$ 
&$\checkmark$
&{83.41$\pm$0.71}
&{78.85$\pm$0.16}

&{69.77$\pm$0.67} 


\\
$\checkmark$
&$\checkmark$
&\textbf{88.75$\pm$0.49}
&\textbf{80.94$\pm$0.17}
&\textbf{70.95$\pm$0.27}
\\
\bottomrule
\end{tabular}

\label{tab:abstudy}
\end{table}

\subsection{Interpretability Analysis}


There have been increasing concerns about the explainability of pre-trained GNNs, as GNNs can be seen as black boxes.
Thus, we employed S-CGIB to explain the molecule prediction compared to ground-truth explanations on four downstream datasets, i.e., Mutagenicity, BENZENE, Alkane Carbonyl, and Fluoride Carbony \cite{agarwal2023evaluating}.
We used the \textit{Fidelity} score \cite{amara2022graphframex} to assess how well the explanation aligns with S-CGIB, as shown in Table \ref{tab:XGNNs}.
The lower value of the $Fidelity-$ score shows a more reliable explanation, while the higher $Fidelity+$ score implies that more important nodes are recognized.
For S-CGIB, we considered the top 50\% of nodes with the highest attention score as explainable nodes.
For the baseline methods, we identified the top 50\% of nodes with the highest positive saliency values as explainable nodes, following the work \citet{DBLP:conf/cvpr/PopeKRMH19}.
We observed that S-CGIB achieved the highest fidelity scores on almost datasets in terms of the two metrics.
This implies that S-CGIB generates a reliable explanation that is aligned with the model prediction.
Moreover, we conducted the qualitative validation on graph interpretation via visualization on two datasets, i.e., Alkane Carbonyl and Fluoride Carbonyl datasets, as shown in Figure \ref{fig:Qualitative}.
We observed that S-CGIB provided a more accurate interpretation of molecules than the baselines.

\subsection{Model Analysis}

\subsubsection{Ablation Analysis}

We further validate the contribution of graph core $G_c$ and subgraph learning in S-CGIB, as shown in Table \ref{tab:abstudy}.
To validate the importance of $G_c$, we considered the presence when using compression or not.
For the subgraph learning, we evaluated the presence of attention coefficients to explore the significant subgraphs.
We observed that:
(i) S-CGIB with both modules performs best in all the downstream datasets. 
This result indicates that exploring graph core and significant subgraphs is crucial to fully generating a pre-trained model as well as the property prediction in downstream datasets.
(ii) While considering the graph core is important, exploring significant subgraphs is more beneficial for molecular property prediction.
For example, the use of graph core can achieve a second ahead in the BBBP dataset, while the model performance for only exploring significant subgraphs remains a close second ahead of other datasets.
We argue that for molecular property prediction, while both graph core and significant subgraphs are important, capturing only significant subgraphs is slightly more beneficial than considering graph core for molecular prediction in specific downstream tasks.

\subsubsection{Sensitivity Analysis}

We further conducted sensitivity analyses on the choice of graph encoders (Appendix D.3), the subgraph sizes (Appendix D.4), and the number of GIN layers (Appendix D.5).
We observed that the GIN encoder showed the best performance among graph encoders, e.g., GCN, GraphSage, and GT \cite{dwivedi2020generalization}, which matches the previous findings \cite{DBLP:conf/nips/LuongS23}.
For the subgraph size, S-CGIB gained the best performance when the subgraph size was small ($k \le 3$). 
This indicates that the subgraph size should be large enough to capture sufficient information but not larger, which can obtain noisy information.
For the number of GIN layers, the model performance remained stable at $l\ge 3$ in most datasets, which helps S-CGIB capture the global graph structures.

\section{Conclusion}

In this paper, we present a novel pre-training strategy for molecules, named S-CGIB, which can discover graph core and significant subgraphs to generate well-distinguished representations.
The main idea is to explore the graph cores of molecules that contain compressed and sufficient information regarding the reconstruction task conditioned on significant subgraphs under the S-CGIB principle.
By doing so, S-CGIB can generate robust representations, improving performance in molecular property prediction tasks.
The experiments over numerous domains showed that S-CGIB consistently outperforms baselines in various downstream tasks.
Furthermore, S-CGIB also delivers model interpretability regarding the functional group detection task despite being learned from self-supervised learning.

\section{Acknowledgments}

This work was supported by the National Research Foundation of Korea (NRF) grant funded by the Korea government (MSIT) (No. 2022R1F1A1065516 and No. 2022K1A3A1A79089461) (O.-J.L.).




\bibliography{aaai25}

\begin{thebibliography}{51}
\providecommand{\natexlab}[1]{#1}

\bibitem[{Agarwal et~al.(2023)Agarwal, Queen, Lakkaraju, and Zitnik}]{agarwal2023evaluating}
Agarwal, C.; Queen, O.; Lakkaraju, H.; and Zitnik, M. 2023.
\newblock Evaluating explainability for graph neural networks.
\newblock \emph{Scientific Data}, 10(1): 144.

\bibitem[{Alemi et~al.(2017)Alemi, Fischer, Dillon, and Murphy}]{DBLP:conf/iclr/AlemiFD017}
Alemi, A.~A.; Fischer, I.; Dillon, J.~V.; and Murphy, K. 2017.
\newblock Deep Variational Information Bottleneck.
\newblock In \emph{5th International Conference on Learning Representations ({ICLR} 2017)}. Toulon, France: OpenReview.net.

\bibitem[{Amara et~al.(2022)Amara, Ying, Zhang, Han, Zhao, Shan, Brandes, Schemm, and Zhang}]{amara2022graphframex}
Amara, K.; Ying, Z.; Zhang, Z.; Han, Z.; Zhao, Y.; Shan, Y.; Brandes, U.; Schemm, S.; and Zhang, C. 2022.
\newblock GraphFramEx: Towards Systematic Evaluation of Explainability Methods for Graph Neural Networks.
\newblock In \emph{Proceedings of the 1st Learning on Graphs Conference (LoG 2022)}. Virtual Event.

\bibitem[{Cao, Lu, and Xu(2015)}]{cao2015grarep}
Cao, S.; Lu, W.; and Xu, Q. 2015.
\newblock GraRep: Learning Graph Representations with Global Structural Information.
\newblock In \emph{24th International Conference on Information and Knowledge Management ({CIKM} 2015)}, 891--900. Melbourne, VIC, Australia: {ACM}.

\bibitem[{Chechik and Tishby(2002)}]{DBLP:conf/nips/ChechikT02}
Chechik, G.; and Tishby, N. 2002.
\newblock Extracting Relevant Structures with Side Information.
\newblock In \emph{Proceedings of the 15th Advances in Neural Information Information Processing Systems (NIPS 2002)}, 857--864. Vancouver, British Columbia, Canada: {MIT} Press.

\bibitem[{Degen et~al.(2008)Degen, Wegscheid-Gerlach, Zaliani, and Rarey}]{degen2008art}
Degen, J.; Wegscheid-Gerlach, C.; Zaliani, A.; and Rarey, M. 2008.
\newblock On the art of compiling and using'drug-like'chemical fragment spaces.
\newblock \emph{ChemMedChem}, 3(10): 1503.

\bibitem[{Devlin et~al.(2019)Devlin, Chang, Lee, and Toutanova}]{DBLP:conf/naacl/DevlinCLT19}
Devlin, J.; Chang, M.; Lee, K.; and Toutanova, K. 2019.
\newblock {BERT:} Pre-training of Deep Bidirectional Transformers for Language Understanding.
\newblock In \emph{Proceedings of the North American Chapter of the Association for Computational Linguistics: Human Language Technologies ({NAACL-HLT} 2019)}, 4171--4186. Minneapolis, MN, USA: ACL.

\bibitem[{Dwivedi and Bresson(2021)}]{dwivedi2020generalization}
Dwivedi, V.~P.; and Bresson, X. 2021.
\newblock A Generalization of Transformer Networks to Graphs.
\newblock In \emph{Proceedings of the AAAI Workshop on Deep Learning on Graphs: Methods and Applications (AAAIW 2021)}.

\bibitem[{Dwivedi et~al.(2022)Dwivedi, Ramp{\'{a}}sek, Galkin, Parviz, Wolf, Luu, and Beaini}]{DBLP:conf/nips/DwivediRGPWLB22}
Dwivedi, V.~P.; Ramp{\'{a}}sek, L.; Galkin, M.; Parviz, A.; Wolf, G.; Luu, A.~T.; and Beaini, D. 2022.
\newblock Long Range Graph Benchmark.
\newblock In \emph{Proceedings of the 35th Advances in Neural Information Processing System (NeurIPS 2022)}. New Orleans, LA, USA.

\bibitem[{Fey and Lenssen(2019)}]{FeyLenssen2019}
Fey, M.; and Lenssen, J.~E. 2019.
\newblock Fast Graph Representation Learning with {PyTorch Geometric}.
\newblock In \emph{Proceedings of the ICLR Workshop on Representation Learning on Graphs and Manifolds (ICLRW 2019)}.

\bibitem[{Gilmer et~al.(2017)Gilmer, Schoenholz, Riley, Vinyals, and Dahl}]{DBLP:journals/corr/GilmerSRVD17}
Gilmer, J.; Schoenholz, S.~S.; Riley, P.~F.; Vinyals, O.; and Dahl, G.~E. 2017.
\newblock Neural Message Passing for Quantum Chemistry.
\newblock In \emph{Proceedings of the 34th International Conference on Machine Learning ({ICML} 2017)}, volume~70 of \emph{PMLR}, 1263--1272. Sydney, NSW, Australia: {PMLR}.

\bibitem[{Gondek and Hofmann(2003)}]{gondek2003conditional}
Gondek, D.; and Hofmann, T. 2003.
\newblock Conditional information bottleneck clustering.
\newblock In \emph{Proceedings of the 3rd IEEE international conference on data mining, workshop on clustering large data sets (ICMDW 2003)}, 36--42.

\bibitem[{Hamilton, Ying, and Leskovec(2017{\natexlab{a}})}]{DBLP:journals/corr/HamiltonYL17}
Hamilton, W.~L.; Ying, Z.; and Leskovec, J. 2017{\natexlab{a}}.
\newblock Inductive Representation Learning on Large Graphs.
\newblock In \emph{Proceedings of the 30th Annual Conference on Neural Information Processing Systems (NeurIPS 2017)}, 1024--1034. Long Beach, CA, {USA}.

\bibitem[{Hamilton, Ying, and Leskovec(2017{\natexlab{b}})}]{DBLP:conf/nips/HamiltonYL17}
Hamilton, W.~L.; Ying, Z.; and Leskovec, J. 2017{\natexlab{b}}.
\newblock Inductive Representation Learning on Large Graphs.
\newblock In \emph{Proceedings of the 30th Annual Conference on Neural Information Processing Systems (NeurIPS 2017)}, 1024--1034. Long Beach, CA, {USA}.

\bibitem[{Hao et~al.(2020)Hao, Lu, Huang, Wang, Hu, Liu, Chen, and Lee}]{DBLP:conf/kdd/HaoLHWHLCL20}
Hao, Z.; Lu, C.; Huang, Z.; Wang, H.; Hu, Z.; Liu, Q.; Chen, E.; and Lee, C. 2020.
\newblock {ASGN:} An Active Semi-supervised Graph Neural Network for Molecular Property Prediction.
\newblock In \emph{Proceedings of the 26th {ACM} {SIGKDD} Conference on Knowledge Discovery and Data Mining (KDD 2020)}, 731--752. Virtual Event: {ACM}.

\bibitem[{Hoang et~al.(2023)Hoang, Jeon, You, Yoon, Jung, and Lee}]{s23084168}
Hoang, V.~T.; Jeon, H.-J.; You, E.-S.; Yoon, Y.; Jung, S.; and Lee, O.-J. 2023.
\newblock Graph Representation Learning and Its Applications: A Survey.
\newblock \emph{Sensors}, 23(8).

\bibitem[{Hoang and Lee(2024)}]{DBLP:conf/aaai/HoangL24}
Hoang, V.~T.; and Lee, O. 2024.
\newblock Transitivity-Preserving Graph Representation Learning for Bridging Local Connectivity and Role-Based Similarity.
\newblock In \emph{Proceedings of the 38th Conference on Artificial Intelligence ({AAAI} 2024)}, 12456--12465. Vancouver,Canada: {AAAI} Press.

\bibitem[{Hu et~al.(2021)Hu, Fey, Ren, Nakata, Dong, and Leskovec}]{DBLP:conf/nips/HuFRNDL21}
Hu, W.; Fey, M.; Ren, H.; Nakata, M.; Dong, Y.; and Leskovec, J. 2021.
\newblock {OGB-LSC:} {A} Large-Scale Challenge for Machine Learning on Graphs.
\newblock In \emph{Proceedings of the 1st Neural Information Processing Systems Track on Datasets and Benchmarks (NeurIPS 2021)}. Virtual Event.

\bibitem[{Hu et~al.(2020{\natexlab{a}})Hu, Liu, Gomes, Zitnik, Liang, Pande, and Leskovec}]{DBLP:conf/iclr/HuLGZLPL20}
Hu, W.; Liu, B.; Gomes, J.; Zitnik, M.; Liang, P.; Pande, V.~S.; and Leskovec, J. 2020{\natexlab{a}}.
\newblock Strategies for Pre-training Graph Neural Networks.
\newblock In \emph{Proceedings of the 8th International Conference on Learning Representations ({ICLR} 2020)}. Addis Ababa, Ethiopia: OpenReview.net.

\bibitem[{Hu et~al.(2020{\natexlab{b}})Hu, Dong, Wang, Chang, and Sun}]{DBLP:conf/kdd/HuDWCS20}
Hu, Z.; Dong, Y.; Wang, K.; Chang, K.; and Sun, Y. 2020{\natexlab{b}}.
\newblock {GPT-GNN:} Generative Pre-Training of Graph Neural Networks.
\newblock In \emph{Proceedings of the 26th {SIGKDD} Conference on Knowledge Discovery and Data Mining (KDD 2020)}, 1857--1867. Virtual Event: {ACM}.

\bibitem[{Inae, Liu, and Jiang(2023)}]{inae2024motifaware}
Inae, E.; Liu, G.; and Jiang, M. 2023.
\newblock Motif-aware Attribute Masking for Molecular Graph Pre-training.
\newblock \emph{arXiv preprint}, arXiv:2309.04589.

\bibitem[{Jang, Gu, and Poole(2017)}]{DBLP:conf/iclr/JangGP17}
Jang, E.; Gu, S.; and Poole, B. 2017.
\newblock Categorical Reparameterization with Gumbel-Softmax.
\newblock In \emph{Proceedings of the 5th International Conference on Learning Representations ({ICLR} 2017)}. Toulon, France: OpenReview.net.

\bibitem[{Kazius, McGuire, and Bursi(2005)}]{kazius2005derivation}
Kazius, J.; McGuire, R.; and Bursi, R. 2005.
\newblock Derivation and validation of toxicophores for mutagenicity prediction.
\newblock \emph{Journal of medicinal chemistry}, 48(1): 312--320.

\bibitem[{Kearnes et~al.(2016)Kearnes, McCloskey, Berndl, Pande, and Riley}]{DBLP:journals/jcamd/KearnesMBPR16}
Kearnes, S.; McCloskey, K.; Berndl, M.; Pande, V.~S.; and Riley, P. 2016.
\newblock Molecular graph convolutions: moving beyond fingerprints.
\newblock \emph{Journal of Computer-Aided Molecular Design}, 30(8): 595--608.

\bibitem[{Kipf and Welling(2017)}]{DBLP:journals/corr/KipfW16}
Kipf, T.~N.; and Welling, M. 2017.
\newblock Semi-Supervised Classification with Graph Convolutional Networks.
\newblock In \emph{Proceedings of the 5th International Conference on Learning Representations ({ICLR} 2017)}. Toulon, France: OpenReview.net.

\bibitem[{Kong et~al.(2022)Kong, Huang, Tan, and Liu}]{DBLP:conf/nips/Kong0T022}
Kong, X.; Huang, W.; Tan, Z.; and Liu, Y. 2022.
\newblock Molecule Generation by Principal Subgraph Mining and Assembling.
\newblock In \emph{Proceedings of the 35th Advances in Neural Information Processing Systems (NeurIPS 2022)}. New Orleans, LA, USA.

\bibitem[{Lee, Lee, and Park(2022)}]{DBLP:conf/aaai/Lee0P22}
Lee, N.; Lee, J.; and Park, C. 2022.
\newblock Augmentation-Free Self-Supervised Learning on Graphs.
\newblock In \emph{Proceedings of the 36th Conference on Artificial Intelligence ({AAAI} 2022)}, 7372--7380. Virtual Event: {AAAI} Press.

\bibitem[{Liu et~al.(2022)Liu, Wang, Liu, Lasenby, Guo, and Tang}]{DBLP:conf/iclr/LiuWLLGT22}
Liu, S.; Wang, H.; Liu, W.; Lasenby, J.; Guo, H.; and Tang, J. 2022.
\newblock Pre-training Molecular Graph Representation with 3D Geometry.
\newblock In \emph{Proceedings of the 10th International Conference on Learning Representations ({ICLR} 2022)}. Virtual Event: OpenReview.net.

\bibitem[{Liu et~al.(2023)Liu, Shi, Zhang, Zhang, Kawaguchi, Wang, and Chua}]{DBLP:conf/nips/LiuSZZKWC23}
Liu, Z.; Shi, Y.; Zhang, A.; Zhang, E.; Kawaguchi, K.; Wang, X.; and Chua, T. 2023.
\newblock Rethinking Tokenizer and Decoder in Masked Graph Modeling for Molecules.
\newblock In \emph{Proceedings of the 36th Advances in Neural Information Processing Systems (NeurIPS 2023)}. New Orleans, LA, USA.

\bibitem[{Luong and Singh(2023)}]{DBLP:conf/nips/LuongS23}
Luong, K.; and Singh, A.~K. 2023.
\newblock Fragment-based Pretraining and Finetuning on Molecular Graphs.
\newblock In \emph{Proceedings of the 36th Advances in Neural Information Processing Systems (NeurIPS 2023)}. New Orleans, LA, USA.

\bibitem[{Maddison, Mnih, and Teh(2017)}]{DBLP:conf/iclr/MaddisonMT17}
Maddison, C.~J.; Mnih, A.; and Teh, Y.~W. 2017.
\newblock The Concrete Distribution: A Continuous Relaxation of Discrete Random Variables.
\newblock In \emph{Proceedings of the 5th International Conference on Learning Representations (ICLR 2017)}. Toulon, France.

\bibitem[{Morris et~al.(2020)Morris, Kriege, Bause, Kersting, Mutzel, and Neumann}]{Morris_2020}
Morris, C.; Kriege, N.~M.; Bause, F.; Kersting, K.; Mutzel, P.; and Neumann, M. 2020.
\newblock TUDataset: A collection of benchmark datasets for learning with graphs.
\newblock In \emph{Proceedings of the ICML 2020 Workshop on Graph Representation Learning and Beyond (ICMLW 2020)}.

\bibitem[{Pope et~al.(2019)Pope, Kolouri, Rostami, Martin, and Hoffmann}]{DBLP:conf/cvpr/PopeKRMH19}
Pope, P.~E.; Kolouri, S.; Rostami, M.; Martin, C.~E.; and Hoffmann, H. 2019.
\newblock Explainability Methods for Graph Convolutional Neural Networks.
\newblock In \emph{Proceedings of the {IEEE} Conference on Computer Vision and Pattern Recognition ({CVPR} 2019)}, 10772--10781. Long Beach, CA, USA: Computer Vision Foundation / {IEEE}.

\bibitem[{Qiu et~al.(2020)Qiu, Chen, Dong, Zhang, Yang, Ding, Wang, and Tang}]{DBLP:conf/kdd/QiuCDZYDWT20}
Qiu, J.; Chen, Q.; Dong, Y.; Zhang, J.; Yang, H.; Ding, M.; Wang, K.; and Tang, J. 2020.
\newblock {GCC:} Graph Contrastive Coding for Graph Neural Network Pre-Training.
\newblock In \emph{Proceedings of the Conference on Knowledge Discovery and Data Mining (KDD 2020)}, 1150--1160. Virtual Event: {ACM}.

\bibitem[{Rong et~al.(2020)Rong, Bian, Xu, Xie, Wei, Huang, and Huang}]{DBLP:conf/nips/RongBXX0HH20}
Rong, Y.; Bian, Y.; Xu, T.; Xie, W.; Wei, Y.; Huang, W.; and Huang, J. 2020.
\newblock Self-Supervised Graph Transformer on Large-Scale Molecular Data.
\newblock In \emph{Proceedings of the 34th Annual Conference on Neural Information Processing Systems (NeurIPS 2020)}. Virtual Event.

\bibitem[{St{\"{a}}rk et~al.(2022)St{\"{a}}rk, Beaini, Corso, Tossou, Dallago, G{\"{u}}nnemann, and Li{\'{o}}}]{DBLP:conf/icml/StarkBCTDGL22}
St{\"{a}}rk, H.; Beaini, D.; Corso, G.; Tossou, P.; Dallago, C.; G{\"{u}}nnemann, S.; and Li{\'{o}}, P. 2022.
\newblock 3D Infomax improves GNNs for Molecular Property Prediction.
\newblock In \emph{Proceedings of the International Conference on Machine Learning ({ICML} 2022)}, volume 162 of \emph{PMLR}, 20479--20502. Baltimore, Maryland, {USA}: {PMLR}.

\bibitem[{Sterling and Irwin(2015)}]{sterling2015zinc}
Sterling, T.; and Irwin, J.~J. 2015.
\newblock ZINC 15--ligand discovery for everyone.
\newblock \emph{Journal of chemical information and modeling}, 55(11): 2324--2337.

\bibitem[{Subramonian(2021)}]{DBLP:conf/aaai/Subramonian21}
Subramonian, A. 2021.
\newblock MOTIF-Driven Contrastive Learning of Graph Representations.
\newblock In \emph{Proceedings of the 35th Conference on Artificial Intelligence ({AAAI} 2021)}, 15980--15981. Virtual Event: {AAAI} Press.

\bibitem[{Tishby, Pereira, and Bialek(2000)}]{DBLP:journals/corr/physics-0004057}
Tishby, N.; Pereira, F. C.~N.; and Bialek, W. 2000.
\newblock The information bottleneck method.
\newblock \emph{arXiv preprint}, arXiv:physics-0004057.

\bibitem[{Velickovic et~al.(2018)Velickovic, Cucurull, Casanova, Romero, Li{\`{o}}, and Bengio}]{velivckovic2017graph}
Velickovic, P.; Cucurull, G.; Casanova, A.; Romero, A.; Li{\`{o}}, P.; and Bengio, Y. 2018.
\newblock Graph Attention Networks.
\newblock In \emph{Proceedings of the 6th International Conference on Learning Representations ({ICLR} 2018)}. Vancouver, BC, Canada: OpenReview.net.

\bibitem[{Velickovic et~al.(2019)Velickovic, Fedus, Hamilton, Li{\`{o}}, Bengio, and Hjelm}]{velickovic2019deep}
Velickovic, P.; Fedus, W.; Hamilton, W.~L.; Li{\`{o}}, P.; Bengio, Y.; and Hjelm, R.~D. 2019.
\newblock Deep Graph Infomax.
\newblock In \emph{Proceedings of the 7th International Conference on Learning Representations ({ICLR} 2019)}. New Orleans, LA, USA: OpenReview.net.

\bibitem[{Wang et~al.(2019)Wang, Zheng, Ye, Gan, Li, Song, Zhou, Ma, Yu, Gai, Xiao, He, Karypis, Li, and Zhang}]{wang2019dgl}
Wang, M.; Zheng, D.; Ye, Z.; Gan, Q.; Li, M.; Song, X.; Zhou, J.; Ma, C.; Yu, L.; Gai, Y.; Xiao, T.; He, T.; Karypis, G.; Li, J.; and Zhang, Z. 2019.
\newblock Deep Graph Library: A Graph-Centric, Highly-Performant Package for Graph Neural Networks.
\newblock \emph{arXiv preprint arXiv:1909.01315}.

\bibitem[{Wu et~al.(2018)Wu, Ramsundar, Feinberg, Gomes, Geniesse, Pappu, Leswing, and Pande}]{wu2018moleculenet}
Wu, Z.; Ramsundar, B.; Feinberg, E.~N.; Gomes, J.; Geniesse, C.; Pappu, A.~S.; Leswing, K.; and Pande, V. 2018.
\newblock MoleculeNet: a benchmark for molecular machine learning.
\newblock \emph{Chemical science}, 9(2): 513--530.

\bibitem[{Xu et~al.(2019)Xu, Hu, Leskovec, and Jegelka}]{DBLP:conf/iclr/XuHLJ19}
Xu, K.; Hu, W.; Leskovec, J.; and Jegelka, S. 2019.
\newblock How Powerful are Graph Neural Networks?
\newblock In \emph{Proceedings of the 7th International Conference on Learning Representations ({ICLR} 2019)}. New Orleans, LA, USA: OpenReview.net.

\bibitem[{Xu et~al.(2021)Xu, Wang, Ni, Guo, and Tang}]{DBLP:conf/icml/XuWNGT21}
Xu, M.; Wang, H.; Ni, B.; Guo, H.; and Tang, J. 2021.
\newblock Self-supervised Graph-level Representation Learning with Local and Global Structure.
\newblock In \emph{Proceedings of the 38th International Conference on Machine Learning ({ICML} 2021)}, volume 139 of \emph{PMLR}, 11548--11558. Virtual Event: {PMLR}.

\bibitem[{You et~al.(2021)You, Chen, Shen, and Wang}]{DBLP:conf/icml/YouCSW21}
You, Y.; Chen, T.; Shen, Y.; and Wang, Z. 2021.
\newblock Graph Contrastive Learning Automated.
\newblock In \emph{Proceedings of the 38th International Conference on Machine Learning ({ICML} 2021)}, volume 139 of \emph{PMLR}, 12121--12132. Virtual Event: {PMLR}.

\bibitem[{You et~al.(2020)You, Chen, Sui, Chen, Wang, and Shen}]{DBLP:conf/nips/YouCSCWS20}
You, Y.; Chen, T.; Sui, Y.; Chen, T.; Wang, Z.; and Shen, Y. 2020.
\newblock Graph Contrastive Learning with Augmentations.
\newblock In \emph{Proceedings of the 33rd Annual Conference on Neural Information Processing Systems (NeurIPS 2020)}. Virtual Event.

\bibitem[{Yu, Cao, and He(2022)}]{DBLP:conf/cvpr/Yu0H22}
Yu, J.; Cao, J.; and He, R. 2022.
\newblock Improving Subgraph Recognition with Variational Graph Information Bottleneck.
\newblock In \emph{Proceedings of the {IEEE/CVF} Conference on Computer Vision and Pattern Recognition ({CVPR} 2022)}, 19374--19383. New Orleans, LA, USA: {IEEE}.

\bibitem[{Yu et~al.(2021)Yu, Xu, Rong, Bian, Huang, and He}]{DBLP:conf/iclr/YuXRBHH21}
Yu, J.; Xu, T.; Rong, Y.; Bian, Y.; Huang, J.; and He, R. 2021.
\newblock Graph Information Bottleneck for Subgraph Recognition.
\newblock In \emph{Proceedings of the 9th International Conference on Learning Representations ({ICLR} 2021)}. Virtual Event: OpenReview.net.

\bibitem[{Zhang et~al.(2020)Zhang, Zhang, Xia, and Sun}]{zhang2020graph}
Zhang, J.; Zhang, H.; Xia, C.; and Sun, L. 2020.
\newblock Graph-Bert: Only Attention is Needed for Learning Graph Representations.
\newblock \emph{CoRR}, abs/2001.05140.

\bibitem[{Zhang et~al.(2021)Zhang, Liu, Wang, Lu, and Lee}]{DBLP:conf/nips/ZhangLWLL21}
Zhang, Z.; Liu, Q.; Wang, H.; Lu, C.; and Lee, C. 2021.
\newblock Motif-based Graph Self-Supervised Learning for Molecular Property Prediction.
\newblock In \emph{Proceedings of the 34th Advances in Neural Information Processing Systems (NeurIPS 2021)}, 15870--15882. Virtual Event.

\end{thebibliography}

\newpage 

$~$  

\newpage



\section{Appendix}

\subsection{A. Proofs}
\subsubsection{A.1. Proof of Equation 15 in the main text}$\\$
\label{sec:app:A_1}
Recall that Eq. 15 in the main text is given as:
\begin{eqnarray}
\label{eq:min12}
& -I({G};{{G}_{c}},{{S}})\le {\mathbb{E}_{{G};{{G}_{c}},{{S}}}} [ -\log {{p}_{\zeta}}( {G}|{{G}_{c}},{{S}} ) ] .   \
\end{eqnarray}%

By employing variational approximation $p_{\zeta} ( {G} | G_c ,S)$ to approximate intractable distribution $p ( {G} | G_c ,S)$, we can have:
\begin{align}
 I\left( G;{{G}_{c}},S \right)&={\mathbb{E}_{G,{{G}_{c}},S}}\left[ \log \frac{p\left( G|{{G}_{c}},S \right)}{p\left( G \right)} \right] \nonumber \\ 
& ={\mathbb{E}_{G,{{G}_{c}},S}}\left[ \log \frac{{{p}_{\zeta }}\left( G|{{G}_{c}},S \right)}{p\left( G \right)} \right] \\ 
& +{\mathbb{E}_{{{G}_{c}},S}}\left[ KL\left( p\left( G|{{G}_{c}},S \right)||{{p}_{\zeta }}\left( G|{{G}_{c}},S \right) \right) \right]. \ \nonumber
\end{align}

Based on the Non-negativity of Kullback–Leibler Divergence, we have:
\begin{align}
I\left( G;{{G}_{c}},S \right)&  \ge {\mathbb{E}_{G,{{G}_{c}},S}}\left[ \log \frac{{{p}_{\zeta }}\left( G|{{G}_{c}},S \right)}{p\left( G \right)} \right] \\ 
& ={\mathbb{E}_{G,{{G}_{c}},S}}\left[ \log{{p}_{\zeta }}\left( G|{{G}_{c}},S \right) \right]+H\left( G \right).  \nonumber
\end{align}

\subsubsection{A.2. Proof of Equation 17 in the main text} $\\$
We first employ a pooling function to get the representation of the graph core as $\Bar{Z} = \text{POOL}(Z)$.
We employ the sufficient encoder assumption that the latent representation $\Bar{Z}$ is lossless in the encoding process, meaning that $\Bar{Z}$ captures all the information from the graph core, i.e., $I( \Bar{Z}| G ) \approx I({{G}_{c}}|{G})$.
We then introduce the upper bound of $I(\Bar{Z}; G)$ by employing a variation approximation, as:
\begin{align}
 I\left( \Bar{Z};G \right)&={\mathbb{E}_{\Bar{Z},G }}\left[ \log \frac{{{p} }\left( \Bar{Z}|G \right)}{p\left( \Bar{Z} \right)} \right] \nonumber \\ 
& ={\mathbb{E}_{ G }} \left[ KL\left( p \left( \Bar{Z} |G \right)||q\left( \Bar{Z} \right) \right) \right]  \\ 
& -{\mathbb{E}_{\Bar{Z},G }} \left[ KL\left( p\left( \Bar{Z} \right)||q\left( \Bar{Z} \right) \right) \right]. \nonumber
\end{align}

Based on the Non-negativity of Kullback–Leibler Divergence, we have:
\begin{align}
\label{eq:11}
I\left( \Bar{Z};G \right)\le {\mathbb{E}_{G}}\left[ KL\left( {{p}_{\phi }}\left( \Bar{Z}|G \right)|| q\left( \Bar{Z} \right) \right) \right].  
\end{align}

Based on the VIB principle \cite{DBLP:conf/iclr/AlemiFD017}, we assume that $q(\Bar{Z})$ can be obtained by aggregating all the node embeddings from $G_c$ in fully perturbed data.
The noise $\epsilon \sim \mathcal{N}(\mu_H, \sigma_H)$ be sampled from the Gaussian distribution, where $\mu_H$ and $\sigma_H$ refer to mean and variance of $H$.
As the summation of a Gaussian distribution results in a Gaussian distribution, we choose sum pooling as the POOLING function; we have:
\begin{align}
\label{eq:22}
q\left( \Bar{Z} \right)=\mathcal{N}\left( N{{\mu }_{H}},N{{\sigma }^2_{H}} \right), 
\end{align}
where $N$ is the number of nodes in the input graph.
Then, for the ${{p}_{\phi }}\left( \Bar{Z}|G  \right)$, we have:

\begin{equation}
\label{eq:33}
\begin{aligned}
p\left( \Bar{Z} |G \right)= & \mathcal{N}\left(  N{{\mu }_{H}}+\sum\limits_{j=1}^{N}{{{\lambda }_{j}}{{H}_{j}}-}\sum\limits_{j=1}^{N}{{{\lambda }_{j}}{{\mu }_{H}},}  \right.\\
&  \left. \sum\limits_{j=1}^{N}{{{\left( 1-{{\lambda }_{j}} \right)}^{2}}\sigma _{H}^{2}}  \right) .
\end{aligned}
\end{equation}
We then plug Eq. \ref{eq:33} and Eq. \ref{eq:22} into Eq. \ref{eq:11}, we have:
\begin{align}
I\left( \Bar{Z};G  \right)\le {\mathbb{E}_{G }}\left[ -\frac{1}{2}\log P+\frac{1}{2N}P+\frac{1}{2N}{{Q}^{2}} \right] + r,
\end{align}
where ${{P} }=\sum\nolimits_{j=1}^{N}{{{\left( 1-{{\lambda }_{j}} \right)}^{2}}}$,
${{Q} }=\frac{\sum\nolimits_{j=1}^{N}{{{\lambda }_{j}}{{\left( {{H}_{j}}-{{\mu }_{H}} \right)}}}}{{{\sigma }_{H}}} $, $r$ denotes a constant value that can be overlooked during the mode training.

\subsection{B. Statistics of Datasets} 
We provide details on the datasets used in our experiments.
For pre-training datasets, we sampled 300k molecules from three datasets, i.e., PCQ4Mv2, QM9 \cite{DBLP:conf/nips/HuFRNDL21}, and mol-PCBA \cite{wu2018moleculenet}).
Specifically, we sampled 100k molecules from each dataset.
The detailed statistics for fine-tuning datasets are summarized in Table \ref{tab:Benchmarks}.
For the datasets used in the interpretability analysis task, we used 4 datasets, e.g., Mutagenicity, BENZENE, Alkane Carbonyl, and Fluoride Carbony, with their ground-truth explanations \cite{agarwal2023evaluating}, as: 
\begin{itemize}
    \item The Mutagenicity (mutag) dataset consists of 1,768 molecules, each labeled according to its mutagenic properties. The dataset is pruned from the original mutagenicity dataset \cite{kazius2005derivation} (4,337 molecules) for the explainability task. 
    \item  The BENZENE dataset consists of 12,000 molecules extracted from the ZINC dataset \cite{sterling2015zinc}, where each molecule is labeled into one of two classes based on the presence or absence of a benzene zing.
    \item  The Alkane Carbonyl dataset consists of 1,125 molecules, each labeled based on the presence of an unbranched alkane and a carbonyl ($C=O$).
    \item  The Fluoride Carbonyl consists of 8,671 molecules, labeled according to the presence of fluoride ($F$-) and a carbonyl ($C=O$).
\end{itemize}

\begin{table*}[tb]
\centering
\caption{The summary of statistics of datasets.}
\setlength{\tabcolsep}{3pt}
\begin{tabular}{l c c c c c c   c}
\toprule
 Category& Dataset&  \# Tasks & Task Type&  \# Graphs& Dimension &  Metric & Avg. \# nodes\\ 
\midrule

\multirow{2}{*}{Biophysics}
& mol-HIV 
& 1 
& Classification
& 41,127
&9
& ROC-AUC
& 25.5
\\


& BACE
& 1
&Classification
&1,513 
& 9
&ROC-AUC
& 34.1
\\

\midrule
\multirow{6}{*}{Physiology}
&BBBP 
&1
&Classification 
&2,039 
& 9
&ROC-AUC
& 23.9
\\

& Tox21 
&12 
&Classification 
&7,831 
& 9
&ROC-AUC
& 18.6
\\

& ToxCast
&617 
&Classification 
&8,575 
& 9
&ROC-AUC
& 18.7
\\

& SIDER 
&27 
&Classification 
&1,427 
& 9
&ROC-AUC
& 33.6
\\

& ClinTox
&2
&Classification
&1,478 
& 9
&ROC-AUC
& 26.1
\\

& MUV
& 17
& Classification
&93,087
& 9
&ROC-AUC
& 24.2
\\


\midrule
\multirow{3}{*}{Physical Chemistry }
& Lipophilicity 
&1 
&Regression 
& 4,200
& 9
&RMSE
&  27.0
\\

& ESOL 
&1 
&Regression 
&1,128 
& 9
&RMSE
& 13.3
\\

& FreeSolv 
&1
&Regression
&642
& 9
&RMSE
& 8.7
\\
\midrule

\multirow{3}{*}{Bioinformatics}
& Mutagenicity
&2
& Classification
&4,337
& 21
&ACC
& 30.32	
\\

&  NCI1
&2
& Classification
&4,110
&37
&ACC
& 29.84
\\
&NCI109 
&2
&Classification
&4,127
&38
&ACC
&29.66
\\

\midrule
 \multirow{2}{*}{LRGB}
& Peptides-func   
&  10
&  Classification
& 15,535
&  9 
&  AP
& 150.94
\\ 

&  Peptides-struct
&  11
&  Regression
&  15,535
&   9 
&  MAE
& 150.94
\\ 
\bottomrule
\end{tabular}
\label{tab:Benchmarks}
\end{table*}

\begin{table*}[tb]
\caption{Performance according to graph encoders.
}
\centering
\setlength{\tabcolsep}{5 pt}
\fontsize{9 pt}{10 pt}\selectfont
\begin{tabular}{l ccc ccc}
\toprule
Encoders
&BBBP
&Tox21
&ToxCast
&SIDER
&MUV
&BACE
\\ 
\midrule
GCN 
&76.53$\pm$0.48 
&68.61$\pm$0.55 
&64.74$\pm$0.35 
&55.54$\pm$0.17 
& {70.14$\pm$0.54} 
& {71.20$\pm$0.02}
\\
GraphSage 
& {83.04$\pm$0.38}
&{77.44$\pm$0.38}
&{67.75$\pm$0.41}
&{57.52$\pm$0.06} 
&67.93$\pm$0.46
&{70.47$\pm$0.36}
\\
GAT
&53.31$\pm$0.61 
&58.50$\pm$1.18 
&57.90$\pm$0.03 
&53.41$\pm$0.50 
&63.92$\pm$0.86 
&57.06$\pm$2.65 
\\
GT 
&{82.55$\pm$1.34}
&{77.75$\pm$0.24} 
&{69.99$\pm$0.37} 
&{57.15$\pm$0.26} 
&{68.27$\pm$0.92}
&68.03$\pm$1.81 
\\
GIN 
&\textbf{88.75$\pm$0.49}
&\textbf{80.94$\pm$0.17}
&\textbf{70.95$\pm$0.27}
&\textbf{64.03$\pm$1.04}
&\textbf{77.71$\pm$1.19} 
&\textbf{86.46$\pm$0.81}
\\
\bottomrule
\end{tabular}
\label{tab:Encoders}
\end{table*}

\subsection{C. Implementation Details}

For the pre-training, we used 5-layer GIN as our GNN encoders for both graph encoding and subgraph learning to learn the representations.
The GIN architecture, which is an expressive power, can distinguish molecular representations with different molecular structures.

\subsubsection{Model training}
Detailed hyperparameter specifications are given in Table \ref{tab:hyperparameter}.
We conducted a search on the embedding dimensions $\{32, 64, 128, 256\}$.
The $\beta$ is determined with a grid search among $\{0.01, 0.1, 1.0, 10 \}$.
For the pre-training phase, we train the model for 600 epochs using Adam optimizer with $1\times 10^{-4}$ learning rate and $1\times 10^{-5}$ weight decay.
We also tune the temperature parameter $\tau$ among $\{1.0, 0.5, 0.1\}$.
\subsubsection{Baselines}
We compared S-CGIB to three groups of baselines, including node-level pre-training, contrastive learning, and subgraph-based pre-training strategies.
These strategies are recent self-supervised pre-training methods on molecular graphs.
We follow closely the settings from these studies for a fair comparison.
For node-level pre-training strategies, we evaluated the S-CGIB performance against three node-level strategies:
\begin{itemize}
    \item  {ContextPred} \cite{DBLP:conf/iclr/HuLGZLPL20} strategy aims at predicting the $k$-hop surrounding structures given a target node.
    \item  {AttrMasking} \cite{DBLP:conf/iclr/HuLGZLPL20} strategy masks the initial feature of node/edge and predicts these features as a node feature recovery task.
    \item  {EdgePred} \cite{DBLP:journals/corr/HamiltonYL17} strategy predicts the presence of an edge between two nodes in graphs.
\end{itemize}
For contrastive learning, we evaluated the S-CGIB performance against five methods:
\begin{itemize}
    \item  {Infomax} \cite{velickovic2019deep} strategy aims to maximize the agreement between the graph-level representations and its sampled subgraphs.
    \item  {JOAO} \cite{DBLP:conf/icml/YouCSW21} strategy generates a set of augmentation schemes, i.e., node dropping, subgraph augmentation, edge perturbation, feature masking, and identical, and try to automatically find the useful augmentations.
    \item  {JOAOv2} \cite{DBLP:conf/icml/YouCSW21} strategy is an improved version of JOAO that can estimate the distributions of initial node features and augmented features to generate more robust augmentation by modifying the projection head.
    \item  {GraphCL} \cite{DBLP:conf/nips/YouCSCWS20} aims at generating multiple views of graphs based on four graph augmentations, i.e., node dropping, edge perturbation, node feature masking, and sampled subgraph, and then maximizes the agreement between these views based on contrastive objectives.
    \item  {GraphLoG} \cite{DBLP:conf/icml/XuWNGT21} discover the global graph structures by using hierarchical prototypes by contrasting graph pairs in a sampled batch.
\end{itemize}
For subgraph-level pre-training strategies, we considered six methods:
\begin{itemize}
    \item  {GraphFP} \cite{DBLP:conf/nips/LuongS23} strategy aims to decompose input molecules into a set of subgraphs based on a dictionary (a bag of subgraphs based on subgraph frequent mining), which benefits the model capture semantic subgraphs.
    \item  {MICRO-Graph} \cite{DBLP:conf/aaai/Subramonian21} strategy aims to generate a bag of prototypical motifs and automatically learn the important functional group-like motifs.
    \item  {MGSSL} \cite{DBLP:conf/nips/ZhangLWLL21} strategy fragments the input molecules into a bag of functional groups based on an improved algorithm of BRICS by considering ring structures.
    \item  {GROVE} \cite{DBLP:conf/nips/RongBXX0HH20} extract 85 functional groups based on discovering frequent subgraphs based on RDKit and utilize a graph transformer architecture.
    \item  {SimSGT} \cite{DBLP:conf/nips/LiuSZZKWC23} employs a set of strategies: breaking the input molecule into smaller subgraphs based on frequent subgraph mining, masking node features, and then recovering these features.
    \item  {MoAMa} \cite{inae2024motifaware} masks the whole sampled motifs and then predicts these features based on a reconstruction task.
\end{itemize}

\begin{table}[tb]
\caption{Hyperparameters of S-CGIB used in experiments.}
\centering
\setlength{\tabcolsep}{6pt}
\begin{tabular}{l c }
\toprule
Hyperparameters
&Values
\\ 
\midrule
Batch size
&128
\\
Number of GIN layers 
&5
\\
Initial feature dimension
&32
\\
Embedding dimension
&64
\\
Number of pre-training epochs
&600
\\
Number of domain adaptation epochs
&50
\\ 
Adam: initial learning rate
&$1\times 10^{-4}$
\\
Adam: weight decay
& $1\times 10^{-5}$
\\
POOLING function
& SUM
\\
$\beta$   
&1.0
\\
$\tau$
&1.0
\\
\bottomrule
\end{tabular}
\label{tab:hyperparameter}
\end{table}

\subsubsection{Training Resources} 
The experiments were conducted in two servers with four NVIDIA RTX A5000 GPUs (24GB RAM/GPU).
Our model was developed and tested in Python 3.8.8 using Torch-geometric \cite{FeyLenssen2019} and DGL Library \cite{wang2019dgl}.
For the environmental settings, we ran the experiments on Ubuntu 20.04 LTS server.

\subsection{D. Additional Experiments} 
\subsubsection{D.1. Performance according to the presence of $I\left(G , S \right )$}

\begin{figure}[tb]
\centering 
  \includegraphics[width=\linewidth]{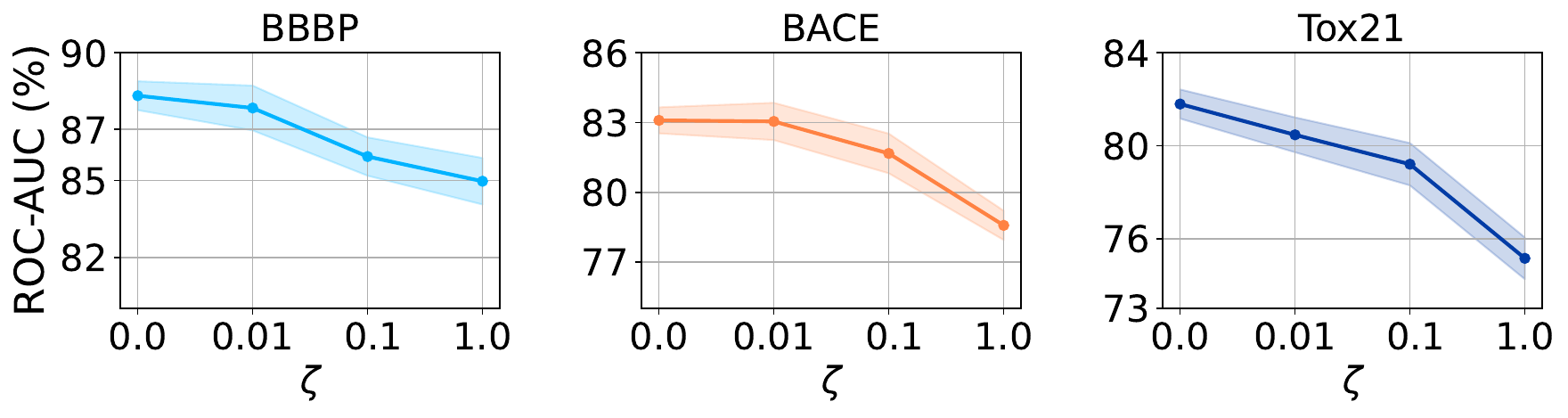}
  \caption{Performance according to weighting factor $\zeta$ for the term $I\left(G ; S \right )$ in Eq. 14.}
  \label{fig:Weight_coefficient}
\end{figure}

Recall that the conditional graph construction term in Eq. 14 is $ \min -I({G};{{G}_{c}}|{{S}})= \min -I({G};{{G}_{c}},{{S}})+I({G};{{S}})$, and our objective is to minimize this term.
Here, we investigate the effect of minimizing the second term, i.e., $I(G; S)$, by conducting experiments on different weight coefficients for this term in the total loss function.
That is, we assign a weight coefficient for the $I(G; S)$ as $\zeta I(G; S)$, along with the total pre-training loss.
Note that when $\zeta = 0$, it indicates that we do not minimize this term.
Formally, we can define the objective as follows:
\begin{align}
\underset{{{f}_{\phi }},{{g}_{\theta }}}{\mathop{\min }}\,\frac{1}{B}\sum\limits_{i=1}^{B}{\frac{1}{N}\sum\limits_{j=1}^{N} {I\left( H_{S}^{j},{\Bar{H}^i} \right)}} , 
\end{align}
where $B$ refers to the number of graphs in a batch,
$N$ is the number of nodes in the input graph $G$,
$\Bar{H} = \text{POOL}(H)$, 
and $H^{j}_S$ is the embeddings of a subgraph rooted at node $j$.
We then can simply model $I(\cdot, \cdot )$ as the dot product, i.e., $I( H_{S}^{j},{\Bar{H}^i} )\approx  H_{S}^{j} \cdot {\Bar{H}^i}$.
As shown in Fig. \ref{fig:Weight_coefficient}, we observed that the model performance decreased when the weight coefficient of $I(G; S)$ increased.
This is because minimizing $-I(G;G_c|S)$ is interpreted in terms of graph structure recovery given $G_c$ conditioned on $S$, i.e., the pre-trained model focuses on the $G_c$ as well as $S$.
We argue that in the learning functional groups, the $S$ are relevant to the graph core $G_c$ to reconstruct the original graph and benefit the model by capturing significant subgraph candidates.

\subsubsection{D.2. Performance according to the use of decoders} 

To validate the use of a decoder and global graph structure preservation, we further conducted experiments with Fully Connected layers as a decoder to reconstruct the input adjacency matrix \cite{DBLP:conf/nips/ZhangLWLL21}. 
Furthermore, we also investigated the reconstruction of a $k$-step transition probability matrix to preserve global graph structures \cite{cao2015grarep}, as shown in Table \ref{tab:abstudy_decoder} ($k $ = 2).
We observed that S-CGIB with two FC layers achieves accuracy comparable to S-CGIB, which indicates that adding FC layers does not necessarily lead to better performance. 
This implies that adjacency matrix reconstruction is sufficient to help S-CGIB generate well-distinguished representations.
Moreover, reconstructing a $k$-step transition probability matrix does not improve the model performance compared to adjacency matrix reconstruction. 
This is because the local neighborhood information is sufficient for the model to learn good representations, achieving a more adequate balance due to the trade-off between compression and reconstruction.

\begin{table}[t]
\caption{
An ablation analysis on different decoders: Fully Connected Layers (FC), Adjacency matrix reconstruction (Adj.), and $k$-step transition matrix reconstruction ($k$-step).
}
\centering
\setlength{\tabcolsep}{ 4.6 pt}
\fontsize{9 pt}{10 pt}\selectfont
\begin{tabular}{ccc ccc }
\toprule
FC
&Adj. 
&$k$-step
&BBBP
&Tox21
&ToxCast
\\ 
\midrule
$\checkmark$ 
&$\checkmark$
&$-$ 
&88.06$\pm$0.37	
&\textbf{81.15$\pm$0.44}
&70.81$\pm$0.63
\\
$\checkmark$ 
&$-$ 
&$\checkmark$
&87.53$\pm$0.42	
&78.63$\pm$0.39
&68.02$\pm$0.81
\\
$-$ 
&$-$ 
&$\checkmark$
&86.14$\pm$0.31	
&78.12$\pm$0.25
&69.04$\pm$0.19
\\
$-$ 
&$\checkmark$
&$-$ 
&\textbf{88.75$\pm$0.49}
&{80.94$\pm$0.17}
&\textbf{70.95$\pm$0.27}
\\

\bottomrule
\end{tabular}

\label{tab:abstudy_decoder}
\end{table}

\begin{figure}[t]
\centering 
  \includegraphics[width=\linewidth]{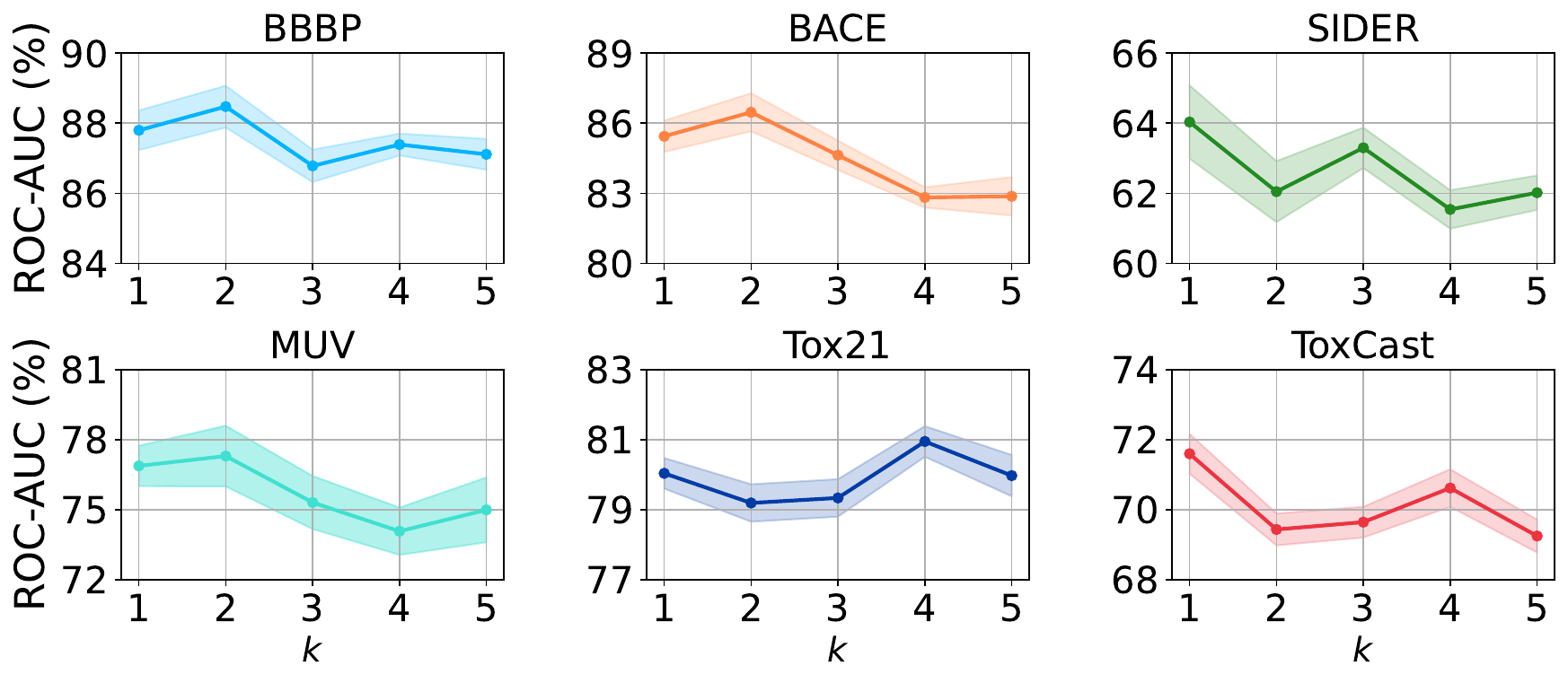}
  \caption{Performance according to subgraph sizes ($k$).
  }
  \label{fig:subgraph_size_k}
\end{figure}

\subsubsection{D.3. Performance according to graph encoders}

Table \ref{tab:Encoders} shows the performances of representative graph encoders, e.g., GCN \cite{DBLP:journals/corr/KipfW16}, GraphSage \cite{DBLP:conf/nips/HamiltonYL17}, GAT \cite{velivckovic2017graph}, and GT \cite{dwivedi2020generalization}, besides GIN \cite{DBLP:conf/iclr/XuHLJ19} encoders.
We observed that the GIN encoder is the most expressive GNN among other graph encoders.
This implies that using an expressive model is crucial to fully utilize
pre-training and that pre-training can even hurt performance when used on models with limited expressive power, e.g., GCN, GraphSAGE, and GAT.
This observation matches with the findings from previous studies \cite{DBLP:conf/iclr/HuLGZLPL20,DBLP:conf/nips/LuongS23,DBLP:conf/iclr/XuHLJ19}.

\subsubsection{D.4. Sensitivity analysis on the subgraph sizes}

We verify the performance of S-CGIB according to the choice of subgraph size $k$, as shown in Figure \ref{fig:subgraph_size_k}.
We observed that when $k$ is small, i.e., 1, 2, 3, the pre-trained model gains the top performance in almost all the datasets.
For example, in the BBBP dataset, the model achieves the highest performance when the size $k=2$ is a 2-hop subgraph rooted at each node.
This implies that the subgraph size should be large enough to capture sufficient and significant subgraphs but not larger, which can obtain noisy nodes.

\subsubsection{D.5. Sensitivity analysis on the number of layers} 

\begin{figure}[t]
\centering 
  \includegraphics[width=\linewidth]{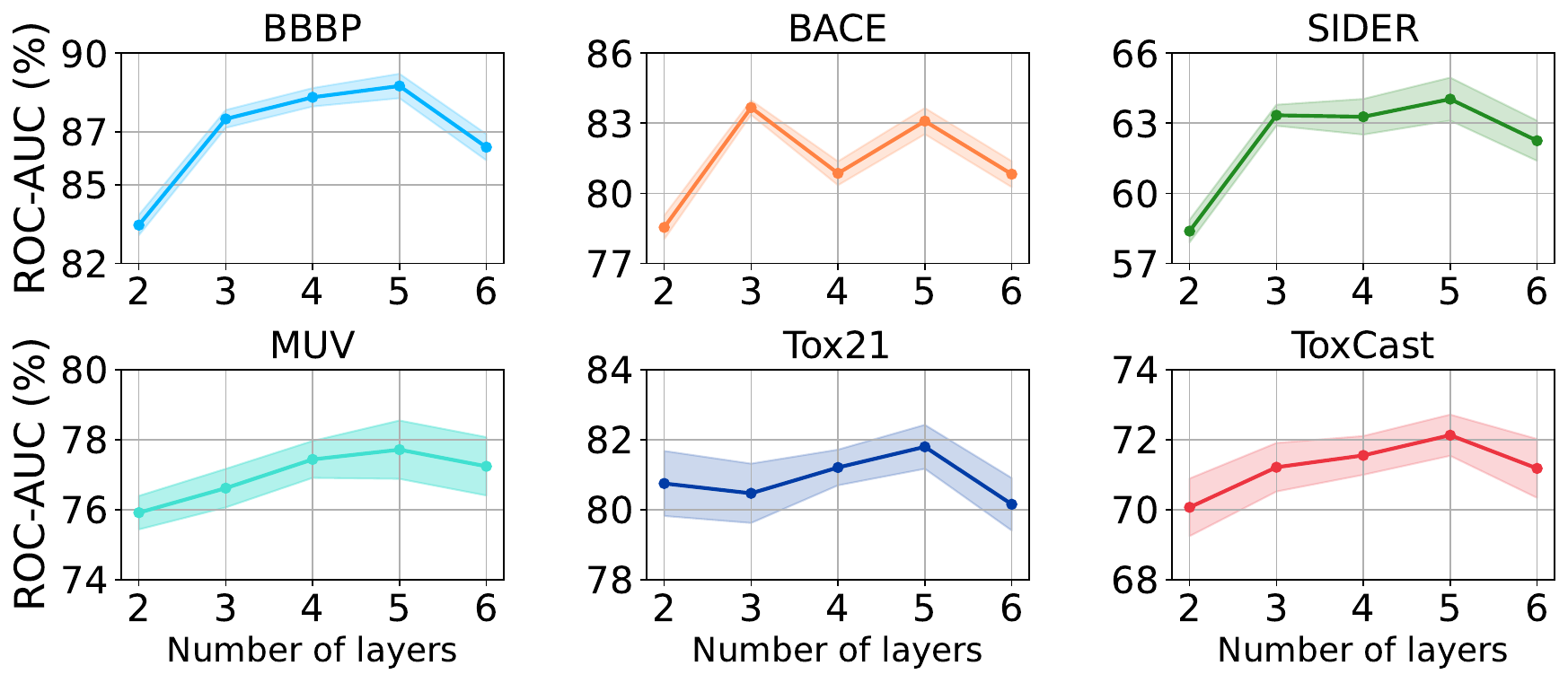}
  \caption{Performance according to the number of GIN layers.}
  \label{fig:size_k}
\end{figure}

We observed that as the number of layers increases, i.e., $l=3$ and $l=5$, the model performance also slightly improved, showing the ability to capture high-order structures of graph core and functional subgraph candidates.

\end{document}